%% file: main.tex
\documentclass[10pt,twocolumn,letterpaper]{article}

\usepackage{cvpr}
\usepackage{times}
\usepackage{epsfig}
\usepackage{graphicx}
\usepackage{amsmath}
\usepackage{amssymb}
\usepackage{makecell}
\usepackage{xfrac}
\usepackage{subcaption}
\usepackage[export]{adjustbox}
\usepackage{color}
\usepackage{rotating}
\usepackage{array}
\usepackage{acronym}

\usepackage{booktabs}
\usepackage{tablefootnote}
\usepackage{colortbl}
\usepackage{tabularx}
\usepackage{multirow}
\usepackage{algorithm,algpseudocode}
\usepackage{pgfplots}
\usepackage{tikz}
\usepackage{pgf-pie}
\usepackage{units}
\usepackage{pifont}
\newcommand{\cmark}{\ding{51}}%
\newcommand{\xmark}{\ding{55}}%
\include{corporateColours}
\renewcommand{\paragraph}[1]{\vspace{8px} \noindent \textbf{#1} \ \ }

\usepackage[pagebackref=true,breaklinks=true,letterpaper=true,colorlinks,bookmarks=false]{hyperref}
\cvprfinalcopy

\ifcvprfinal\fi
\makeatletter
\newenvironment{customlegend}[1][]{%
    \begingroup
    \pgfplots@init@cleared@structures
    \pgfplotsset{#1}%
}{%
    \pgfplots@createlegend
    \endgroup
}%

\def\addlegendimage{\pgfplots@addlegendimage}
\makeatother

\begin{document}

\title{Seeing Through Fog Without Seeing Fog:\\ Deep Multimodal Sensor Fusion in Unseen Adverse Weather \vspace*{-3mm}}

\author{
\resizebox{\linewidth}{!}{
\begin{tabular}{@{}c@{}}
	Mario Bijelic$^{1,3}$\hspace{1.3mm}
	Tobias Gruber$^{1,3}$\hspace{1.3mm}
	Fahim Mannan$^{2}$\hspace{1.3mm}
	Florian Kraus$^{1,3}$\hspace{1.3mm}
	Werner Ritter$^{1}$\hspace{1.3mm}
	Klaus Dietmayer$^{3}$\hspace{1.3mm}
	Felix Heide$^{2,4}$ \vspace{8pt} \\
	\textsuperscript{1}Mercedes-Benz AG\hspace{1.em}
	\textsuperscript{2}Algolux\hspace{1.em}
	\textsuperscript{3}Ulm University\hspace{1.em}
	\textsuperscript{4}Princeton University\hspace{1.em}
\end{tabular}}%
\vspace{-5pt}
}

\maketitle

\vspace{-1em}
\begin{abstract}
	The fusion of multimodal sensor streams, such as camera, lidar, and radar measurements, plays a critical role in object detection for autonomous vehicles, which base their decision making on these inputs. While existing methods exploit redundant information in good environmental conditions, they fail in adverse weather where the sensory streams can be asymmetrically distorted. These rare ``edge-case'' scenarios are not represented in available datasets, and existing fusion architectures are not designed to handle them. To address this challenge we present a novel multimodal dataset acquired in over 10,000~km of driving in northern Europe. Although this dataset is the first large multimodal dataset in adverse weather, with 100k labels for lidar, camera, radar, and gated NIR sensors, it does not facilitate training as extreme weather is rare. To this end, we present a deep fusion network for robust fusion without a large corpus of labeled training data covering all asymmetric distortions. Departing from proposal-level fusion, we propose a single-shot model that adaptively fuses features, driven by measurement entropy. We validate the proposed method, trained on clean data, on our extensive validation dataset. Code and data are available here {\small\url{https://github.com/princeton-computational-imaging/SeeingThroughFog}}.
\end{abstract}

\vspace{-0.7em}
\section{Introduction}
\input{introduction}
\section{Related Work}\label{Sec:relatedwork}
\input{related_work}

\section{Multimodal Adverse Weather Dataset}
\input{dataset}

\section{Adaptive Deep Fusion} 
\input{method}
\section{Assessment}
\input{result}

\section{Conclusion and Future Work}
\input{discussion}
\section*{Acknowledgment}
The authors would like to acknowledge the funding from the European Un\nolinebreak ion\nolinebreak\ un\nolinebreak der\nolinebreak\ the H2020 ECSEL Programme as part of the DENSE project, 
contract number 692449, and thank Jason Taylor for fruitful discussion.
{\small
\bibliographystyle{ieee}
\bibliography{bib}
}
\end{document}

%% file: corporateColours.tex
\definecolor{uni_apfelgruen}{cmyk}{.5, 0, 1, 0}
\definecolor{uni_mittelblau}{cmyk}{1, 0.4, 0, 0}
\definecolor{uni_gelb}{cmyk}{0, 0.1, 1, 0}
\definecolor{uni_rot}{cmyk}{0, 1, 1, 0}
\definecolor{mittelblau}{RGB}{0, 126, 198}
\definecolor{violettblau}{cmyk}{0.9, 0.6, 0, 0}
\definecolor{rot}{RGB}{238, 28 35}
\definecolor{apfelgruen}{RGB}{140, 198, 62}
\definecolor{gelb}{RGB}{255, 229, 0}
\definecolor{orange}{RGB}{244, 111, 33}
\definecolor{pink}{RGB}{237, 0, 140}
\definecolor{lila}{RGB}{128, 10, 145}
\definecolor{hellgrau}{RGB}{224, 224, 224}
\definecolor{mittelgrau}{RGB}{128, 128, 128}
\definecolor{dunkelgrau}{RGB}{80,80,80}
\definecolor{anthrazit}{RGB}{19, 31, 31}
\definecolor{magenta}{RGB}{148,5,132}

\definecolor{conv_blue}{RGB}{80,162,220}
\definecolor{dunkelgruen}{RGB}{0,102,0}
\definecolor{tuerkis}{RGB}{153,204,204}
\definecolor{own_yellow}{RGB}{255,204,0}

\definecolor{dai_ligth_grey}{RGB}{230,230,230}
\definecolor{dai_ligth_grey20K}{RGB}{200,200,200}
\definecolor{dai_ligth_grey40K}{RGB}{158,158,158}
\definecolor{dai_ligth_grey60K}{RGB}{112,112,112}
\definecolor{dai_ligth_grey80K}{RGB}{68,68,68}

\definecolor{dai_petrol}{RGB}{0,103,127}
\definecolor{dai_petrol20K}{RGB}{0,86,106}
\definecolor{dai_petrol40K}{RGB}{0,67,85}
\definecolor{dai_petrol80}{RGB}{0,122,147}
\definecolor{dai_petrol60}{RGB}{80,151,171}
\definecolor{dai_petrol40}{RGB}{121,174,191}
\definecolor{dai_petrol20}{RGB}{166,202,216}

\definecolor{dai_deepred}{RGB}{113,24,12}
\definecolor{dai_deepred20K}{RGB}{90,19,10}
\definecolor{dai_deepred40K}{RGB}{68,14,7}

%% file: introduction.tex
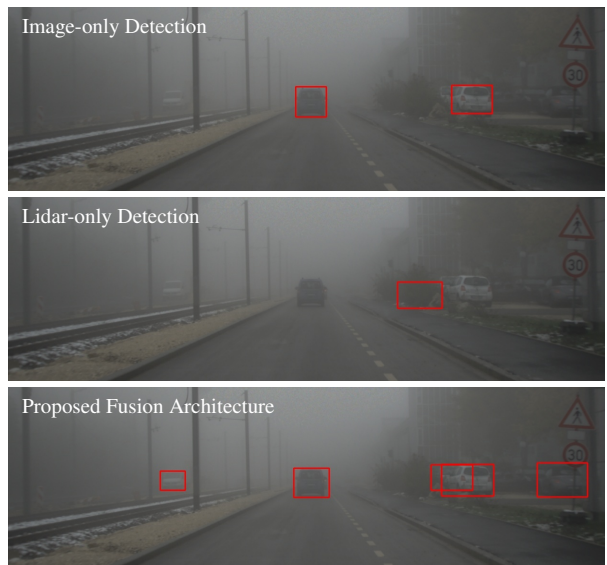
\begin{figure}[t!]
	\centering
	\vspace{-0.5em}
	\hspace{2pt}\begin{tikzpicture}
\node[above right] (img) at (0,0) {\includegraphics[width=0.95\linewidth]{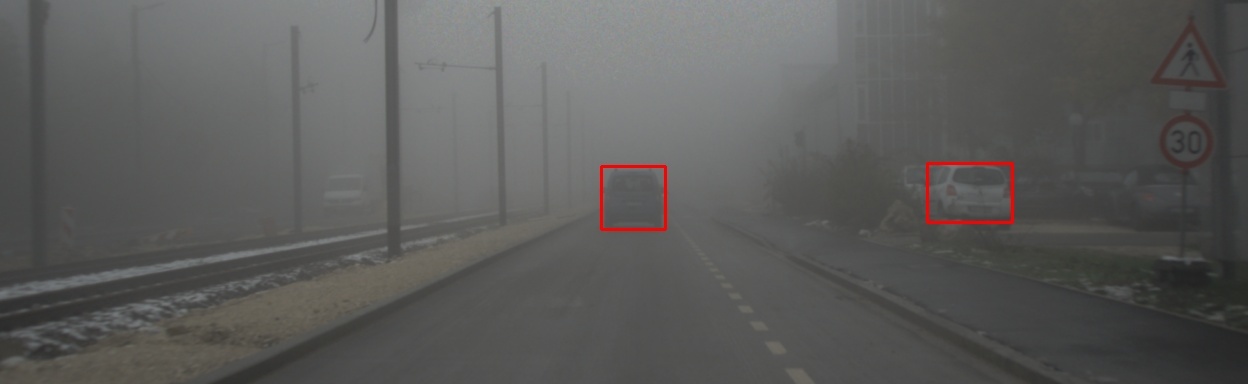}}; 
\node[white, anchor=west] at (5pt,65pt) {{\footnotesize{Image-only Detection}}};
\end{tikzpicture}\\ \vspace{-0.5em}
\hspace{0.2em}\begin{tikzpicture}
\node[above right] (img) at (0,0) {\includegraphics[width=0.95\linewidth]{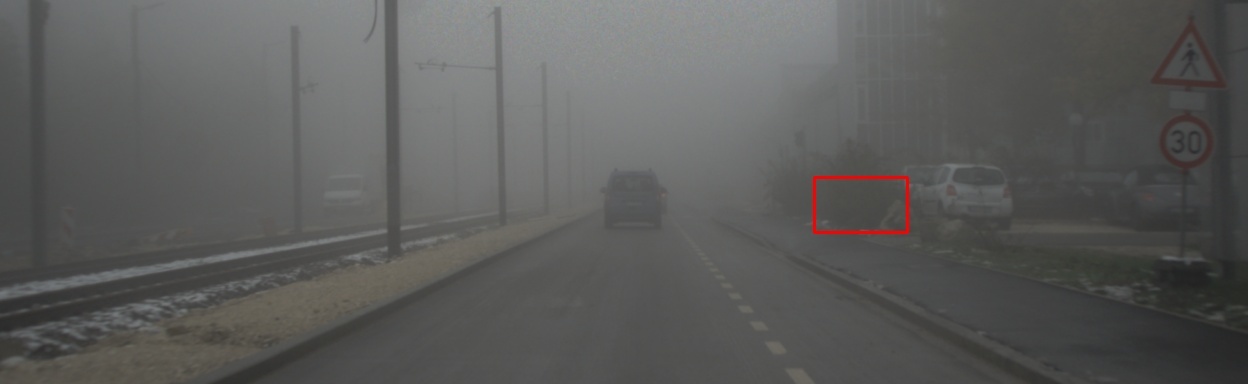}}; 
\node[white, anchor=west] at (5pt,65pt) {{\footnotesize{Lidar-only Detection}}};
\end{tikzpicture}\\ \vspace{-0.5em}
\hspace{0.2em}\begin{tikzpicture}
\node[above right] (img) at (0,0) {\includegraphics[width=0.95\linewidth]{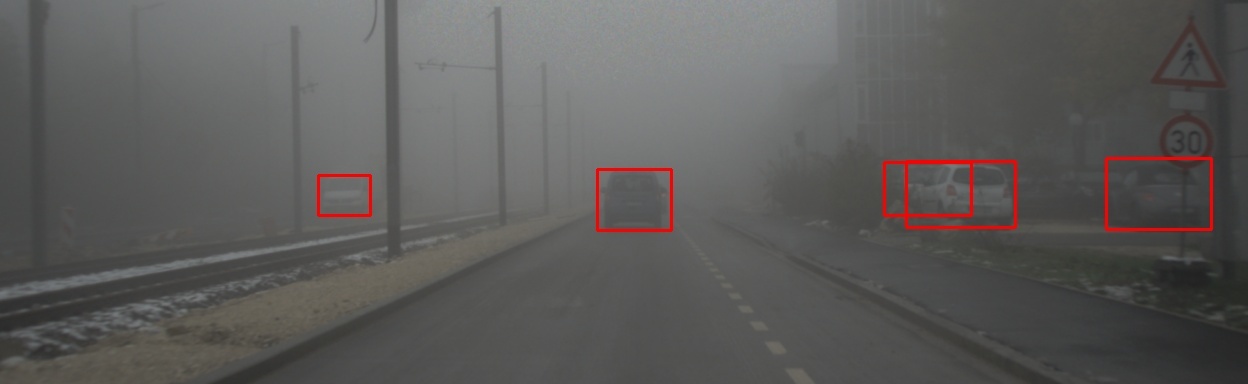}}; 
\node[white, anchor=west] at (5pt,65pt) {{\footnotesize{Proposed Fusion Architecture}}};
\end{tikzpicture}\\ 
    \vspace{-1.0em}   
	\caption{Existing object detection methods, including efficient Single-Shot detectors (SSD)~\cite{SSDLiu2015}, are trained on automotive datasets that are biased towards good weather conditions. While these methods work well in good conditions~\cite{Kitt_dataset,waymo_open_dataset}, they fail in rare weather events (top). Lidar-only detectors, such as the same SSD model trained on projected lidar depth, might be distorted due to severe backscatter in fog or snow (center). These asymmetric distortions are a challenge for fusion methods, that rely on redundant information. The proposed method (bottom) learns to tackle unseen (potentially asymmetric) distortions in multimodal data {without seeing training data} of these rare scenarios.
	}
	\vspace{-1.5em}
	\label{fig:RealWorldPerformance}
\end{figure}

Object detection is a fundamental computer vision problem in autonomous robots, including self-driving vehicles and autonomous drones. Such applications require 2D or 3D bounding boxes of scene objects in challenging real-world scenarios, including complex cluttered scenes, highly varying illumination, and adverse weather conditions. The most promising autonomous vehicle systems rely on redundant inputs from multiple sensor modalities~\cite{waymo_open_dataset, caesar2019nuscenes,BerthaDrive}, including camera, lidar, radar, and emerging sensor such as FIR~\cite{hurney2015review}. A growing body of work on object detection using convolutional neural networks has enabled accurate 2D and 3D box estimation from such multimodal data, typically relying on camera and lidar data~\cite{xu2017pointfusion,chen2017multi,song2016deep,zhou2018voxelnet,yang2018pixor,luo2018fast,ku2018joint}. While these existing methods, and the autonomous system that performs decision making on their outputs, perform well under normal imaging conditions, they fail in adverse weather and imaging conditions. This is because existing training datasets are biased towards clear weather conditions, and
detector architectures are designed to rely only on the redundant information in the undistorted sensory streams. However, they are not designed for harsh scenarios that distort the sensor streams asymmetrically, see Figure.~\ref{fig:RealWorldPerformance}. 
Extreme weather conditions are statistically rare. For example, thick fog is observable only during 0.01\,\% of typical driving in North America, and even in foggy regions, dense fog with visibility below \unit[50]{m} occurs only up to 15 times a year \cite{VanOldenborgh2010a}. Figure~\ref{fig:sweden_histo_weather} shows the distribution of real driving data acquired over four weeks in Sweden covering \unit[10,000]{km} driven in winter conditions. The naturally biased distribution validates that harsh weather scenarios are only rarely or even not at all represented in available datasets~\cite{xu2017end,Kitt_dataset,waymo_open_dataset}. Unfortunately, domain adaptation methods~\cite{murez2018image,Cycada,long2018conditional} also do not offer an ad-hoc solution as they require target samples, and adverse weather-distorted data are underrepresented in general. Moreover, existing methods are limited to image data but not to multisensor data, e.g. including lidar point-cloud data.

Existing fusion methods have been proposed mostly for lidar-camera setups~\cite{xu2017pointfusion,chen2017multi,luo2018fast,ku2018joint, cho2014multi}, as a result of the limited sensor inputs in existing training datasets~\cite{xu2017end,Kitt_dataset,waymo_open_dataset}. These methods do not only struggle with sensor distortions in adverse weather due to the bias of the training data. Either they perform late fusion through filtering after independently processing the individual sensor streams~\cite{cho2014multi}, or they fuse proposals~\cite{ku2018joint} or high-level feature vectors~\cite{xu2017pointfusion}. The network architecture of these approaches is designed with the assumption that the data streams are consistent and redundant, i.e. an object appearing in one sensory stream also appears in the other. However, in harsh weather conditions, such as fog, rain, snow, or extreme lighting condition, including low-light or low-reflectance objects, multimodal sensor configurations can fail asymmetrically. For example, conventional RGB cameras provide unreliable noisy measurements in low-light scene areas, while scanning lidar sensors provide reliable depth using active illumination. In rain and snow, small particles affect the color image and lidar depth estimates equally through backscatter. Adversely, in foggy or snowy conditions, state-of-the-art pulsed lidar systems are restricted to less than \unit[20]{m} range due to backscatter, see Figure~\ref{fig:sensor_performance}. While relying on lidar measurements might be a solution for night driving, it is {not} for adverse weather conditions. 

In this work, we propose a multimodal fusion method for object detection in adverse weather, including fog, snow, and harsh rain, without having large annotated training datasets available for these scenarios. Specifically, we handle asymmetric measurement corruptions in camera, lidar, radar, and gated NIR sensor streams by departing from existing proposal-level fusion methods: we propose an adaptive single-shot deep fusion architecture which exchanges features in intertwined feature extractor blocks. This deep early fusion is steered by measured entropy. The proposed adaptive fusion allows us to learn models that generalize across scenarios. To validate our approach, we address the bias in existing datasets by introducing a novel multimodal dataset acquired on three months of acquisition in northern Europe. This dataset is the first large multimodal driving dataset in adverse weather, with 100k labels for lidar, camera, radar, gated NIR sensor, and FIR sensor. Although the weather-bias still prohibits training, this data allows us to validate that the proposed method generalizes robustly to unseen weather conditions with asymmetric sensor corruptions, while being trained on clean data.

Specifically, we make the following contributions:
\begin{itemize}
	\itemsep-0.4em
	\item We introduce a multimodal adverse weather dataset covering camera, lidar, radar, gated NIR, and FIR sensor data. The dataset contains rare scenarios, such as heavy fog, heavy snow, and severe rain, during more than 10,000~km of driving in northern Europe.
	\item We propose a deep multimodal fusion network which departs from proposal-level fusion, and instead adaptively fuses driven by measurement entropy.
	\item We assess the model on the proposed dataset, validating that it generalizes to unseen asymmetric distortions. The approach outperforms state-of-the-art fusion methods more than 8\%\,AP in hard scenarios independent of weather, including light fog, dense fog, snow, and clear conditions, and it runs in real-time.
\end{itemize}

%

%% file: related_work.tex
\vspace{0.5em}\noindent\textbf{Detection in Adverse Weather Conditions} 
Over the last decade, seminal work on automotive datasets~\cite{brostow2008segmentation,Cordts2016Cityscapes,Kitt_dataset,dollar2012pedestrian,xu2017end,Argoverse} has provided a fertile ground for automotive object detection \cite{chen2017multi,cai2016unified,xu2017pointfusion,ku2018joint,SSDLiu2015,Girshick2015}, depth estimation~\cite{eigen2014depth,liu2016learning,godard2017unsupervised}, lane-detection~\cite{hillel2014recent}, traffic-light detection~\cite{jensen2016vision}, road scene segmentation~\cite{brostow2008segmentation,badrinarayanan2015segnet}, and end-to-end driving models~\cite{bojarski2016end,xu2017end}. Although existing datasets fuel this research area, they are biased towards good weather conditions due to geographic location~\cite{xu2017end} and captured season~\cite{Kitt_dataset}, and thus lack severe distortions introduced by rare fog, severe snow, and rain. A number of recent works explore camera-only approaches in such adverse conditions~\cite{sakaridis2018semantic,Cai2016,Ancuti2018OHAZEAD}. However, these datasets are very small with less than 100 captured images~\cite{sakaridis2018semantic} and limited to camera-only vision tasks. In contrast, existing autonomous driving applications rely on multimodal sensor stacks, including camera, radar, lidar, and emerging sensor, such as gated NIR imaging~\cite{grauer2014active,Gated2Depth}, and have to be evaluated on thousands of hours of driving. 
In this work, we fill this gap and introduce a large scale evaluation set in order to develop a fusion model for such multimodal inputs that is robust to unseen distortions.

\vspace{0.5em}\noindent\textbf{Data Preprocessing in Adverse Weather}
A large body of work explores methods for the removal of sensor distortions before processing. Especially fog and haze removal from conventional intensity image data have been explored extensively~\cite{Yang2018, Zhang2018, Ki2018, Sim2018, Kupyn2017, Cai2016, Li2017, EntropyDehazing}. Fog results in a distance-dependent loss in contrast and color. Fog removal methods have not only been suggested for display application~\cite{he2011single}, it has also been proposed as preprocessing to improve the performance of downstream semantic tasks~\cite{sakaridis2018semantic}.
Existing fog and haze removal methods rely on scene priors on the latent clear image and depth to solve the ill-posed recovery. These priors are either hand-crafted~\cite{he2011single} and used for depth and transmission estimation separately, or they are learned jointly as part of trainable end-to-end models~\cite{Li2017, pix2pix2016, Zhu2017}. Existing methods for fog and visibility estimation~\cite{fogest2014,tarel2010improved} have been proposed for camera driver-assistance systems. Image restoration approaches have also been applied to deraining \cite{Chen2018} or deblurring \cite{Kupyn2017}.

\vspace{0.5em}\noindent\textbf{Domain Adaptation}
Another line of research tackles the shift of unlabeled data distributions by domain adaptation \cite{ADDA, Cycada, Sakaridis2019GuidedCM, Hoffman2014ContinuousMB, Zhang2017CurriculumDA, vu2019dada}. Such methods could be applied to adapt clear labeled scenes to demanding adverse weather scenes \cite{Cycada} or through the adaptation of feature representations \cite{ADDA}. Unfortunately, both of these approaches struggle to generalize, because, in contrast to existing domain transfer methods, weather-distorted data in general, not only labeled data, is underrepresented. Moreover, existing methods do not handle multimodal data.

\vspace{0.5em}\noindent\textbf{Multisensor Fusion}
Multisensor feeds in autonomous vehicles are typically fused to exploit varying cues in the measurements~\cite{mees2016choosing}, simplify path-planning~\cite{dolgov2010path}, to allow for redundancy in the presence of distortions~\cite{premebida2009lidar}, or solve for joint vision tasks, such as 3D object detection~\cite{xu2017pointfusion}. Existing sensing systems for fully-autonomous driving include lidar, camera, and radar sensors. As large automotive datasets~\cite{xu2017end,Kitt_dataset,waymo_open_dataset} cover limited sensory inputs, existing fusion methods have been proposed mostly for lidar-camera setups~\cite{xu2017pointfusion,MVXNet,chen2017multi,ku2018joint,luo2018fast}. Methods such as AVOD~\cite{ku2018joint} and MV3D~\cite{chen2017multi} incorporate multiple views from camera and lidar to detect objects. They rely on the fusion of pooled regions of interest and hence perform late feature fusion following popular region proposal architectures~\cite{Ren2015}. In a different line of research, Qi et al.~\cite{Qi2017} and Xu et al.~\cite{xu2017pointfusion} propose a pipeline model that requires a valid detection output for the camera image and a 3D feature vector extracted from the lidar point-cloud. Kim et al.~\cite{Kim2018} propose a gating mechanism for camera-lidar fusion. In all existing methods, the sensor streams are processed separately in the feature extraction stage, and we show that this prohibits learning redundancies, and, in fact, performs worse than a single sensor stream in the presence of asymmetric measurement distortions.

%% file: dataset.tex
\begin{table}[t]
    \normalsize
    \heavyrulewidth 1.7pt
    \belowrulesep 0pt
    \aboverulesep 0pt
    \centering
\resizebox{1.00\linewidth}{!}{
        \begin{tabular}{@{}l|c|c|c|c|c}
		\toprule
		\textsc{\textbf{Dataset}}       			&   \textbf{KITTI} \cite{Kitt_dataset} & \textbf{BDD} \cite{yu2018bdd100k} & \textbf{Waymo} \cite{waymo_open_dataset} & \textbf{NuScenes} \cite{caesar2019nuscenes} & \textbf{Ours} \\
		\textsc{Sensor Setup}            		   	   	&  & & & & \\
		\toprule
        \textsc{RGB Cameras}  		     	        &  2  & 1   & 5    & 6          & 2  \\
        \textsc{RGB resolution}				        &  1242$\times$372
        &   1280$\times$720 & 1920$\times$1080  & 1600x900  & 1920x1024\\
        \textsc{Lidar Sensors} 				     	&  1  & \xmark    & 5    & 1          & 2  \\
        \textsc{Lidar resolution}			     	&  64 &  0  & 64   & 32         & 64\\
		\textsc{Radar Sensor} 						&  \xmark   & \xmark   & \xmark     & 4 & 1 \\
		\textsc{Gated Camera}						&  \xmark  & \xmark   & \xmark    &  \xmark   & 1 \\
		\textsc{FIR Camera}							&  \xmark  & \xmark   & \xmark   &  \xmark    & 1 \\
		\textsc{Frame rate}							& \unit[10]{Hz} & \unit[30]{Hz} & \unit[10]{Hz} & \unit[1]{Hz}/\unit[10]{Hz} & \unit[10]{Hz} \\
		\midrule
		\textsc{Dataset Statistics}      		   	&     & & & & \\
		\midrule
		\textsc{Labeled Frames}						& 15K    &  100k    &  198k  & 40K  &  13.5K\\
		\textsc{Labels}	     				    & 80k    & 1.47M &  7.87M   & 1.4M &  100K\\
		\textsc{Scene Tags}	     		    & \xmark     &  \cmark   & \xmark & \cmark & \cmark\\
		\textsc{Night Time}	     			& \xmark     &  \cmark   & \cmark & \cmark & \cmark\\
		\textsc{Light Weather}			    & \xmark     &  \cmark   & \xmark & \cmark & \cmark\\
		\textsc{Heavy Weather}			    & \xmark     &  \xmark   & \xmark & \xmark & \cmark\\
		\textsc{Fog Chamber}    & \xmark     &  \xmark   & \xmark & \xmark & \cmark\\
		\bottomrule
		\end{tabular}}
	\vspace{-0.7em}
    \caption{Comparison of the proposed multimodal adverse weather dataset to existing automotive detection datasets.}\label{tab:SOTADatasets} \vspace{-1.5em}
\end{table}
\begin{figure*}[t!]
\vspace{-7mm}
	\centering
	\hspace{2pt}\begin{tikzpicture}
	\node[anchor=west] (img) at (240pt,-9pt) {\includegraphics[width=1.01\columnwidth]{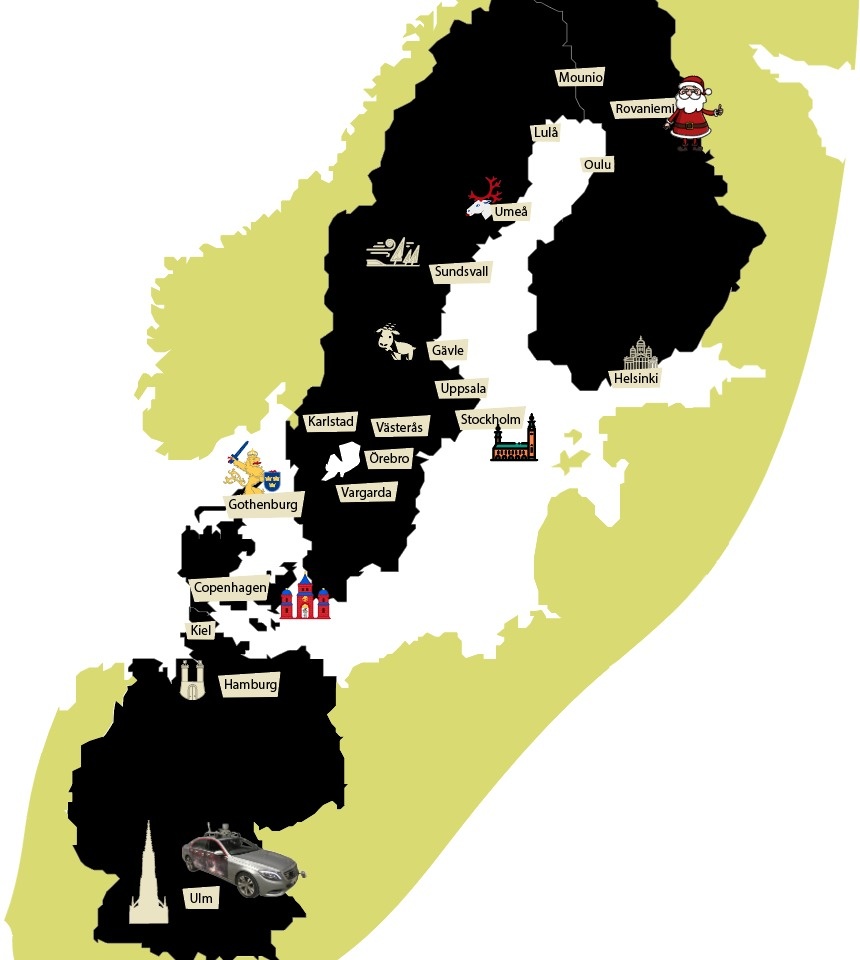}};
	\node[anchor=west] (img) at (-5pt,48pt) {\includegraphics[width=1.1\columnwidth]{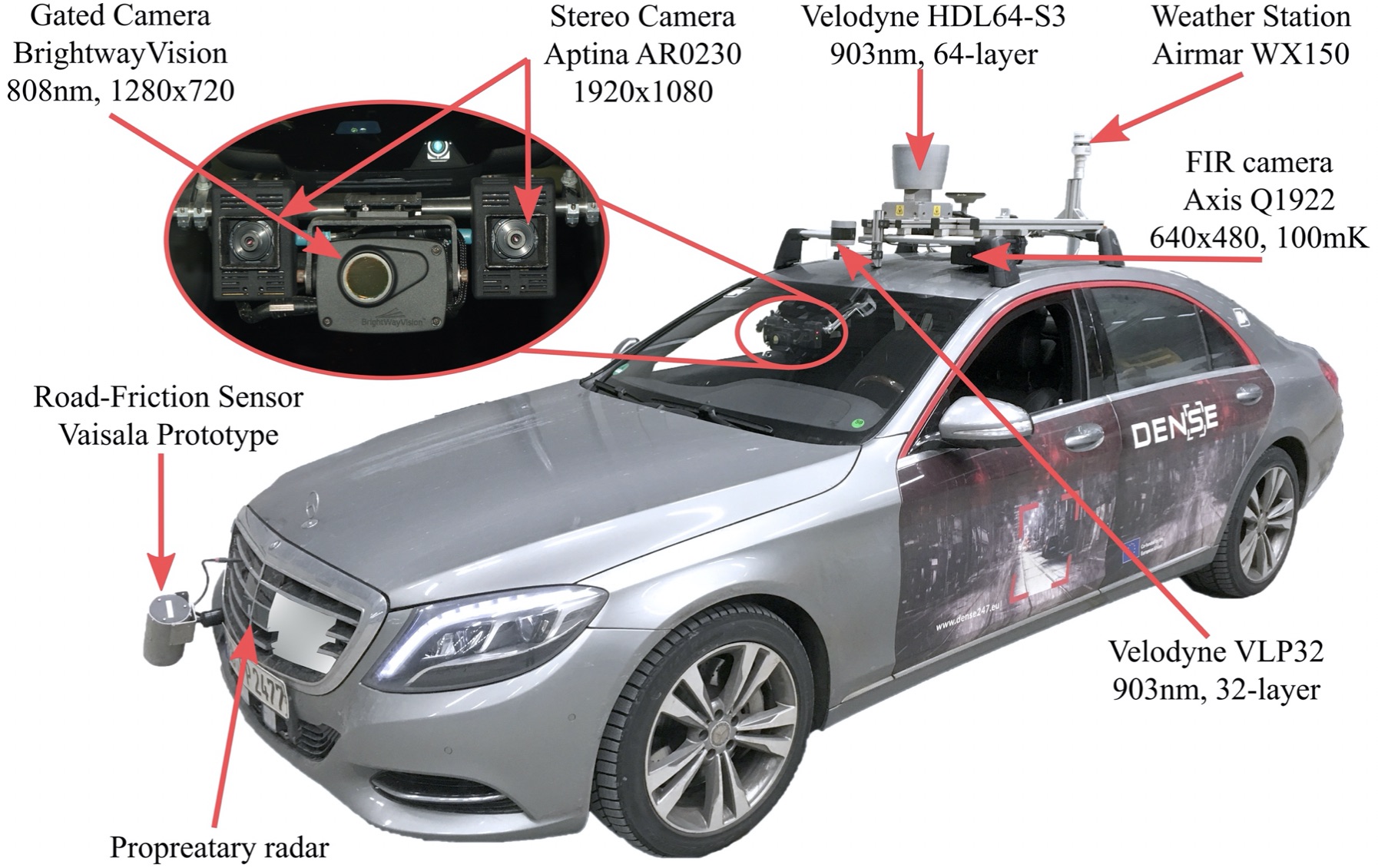}};
	\node[anchor=west] (img) at (5pt,-105pt) {\input{tikz/dataset_sweden_distribution.tikz}};
	\node[black, anchor=west] at (350pt,139 pt) {{{Geographical Sampling}}};
	\node[black, anchor=west] at (120pt,139pt) {{{Vehicle Setup}}};
	\node[black, anchor=west, rotate=90] at (5pt,-130pt) {{\footnotesize{Quantity [\#Frames]}}};
	\end{tikzpicture}
	\vspace{-1.4em}   
	\caption{\textit{Right:} Geographical coverage of the data collection campaign covering two months and \unit[10,000]{km} in Germany, Sweden, Denmark, and Finland. \textit{Top Left:} Test vehicle setup with top-mounted lidar, gated camera with flash illumination, RGB camera, proprietary radar, FIR camera, weather station, and road friction sensor. \textit{Bottom Left:} Distribution of weather conditions throughout the data acquisition. The driving data is highly unbalanced with respect to weather conditions and only contains adverse conditions as rare samples.}
	\vspace*{-1em}
	\label{fig:car}
	\label{fig:sweden_histo_weather}
\end{figure*}
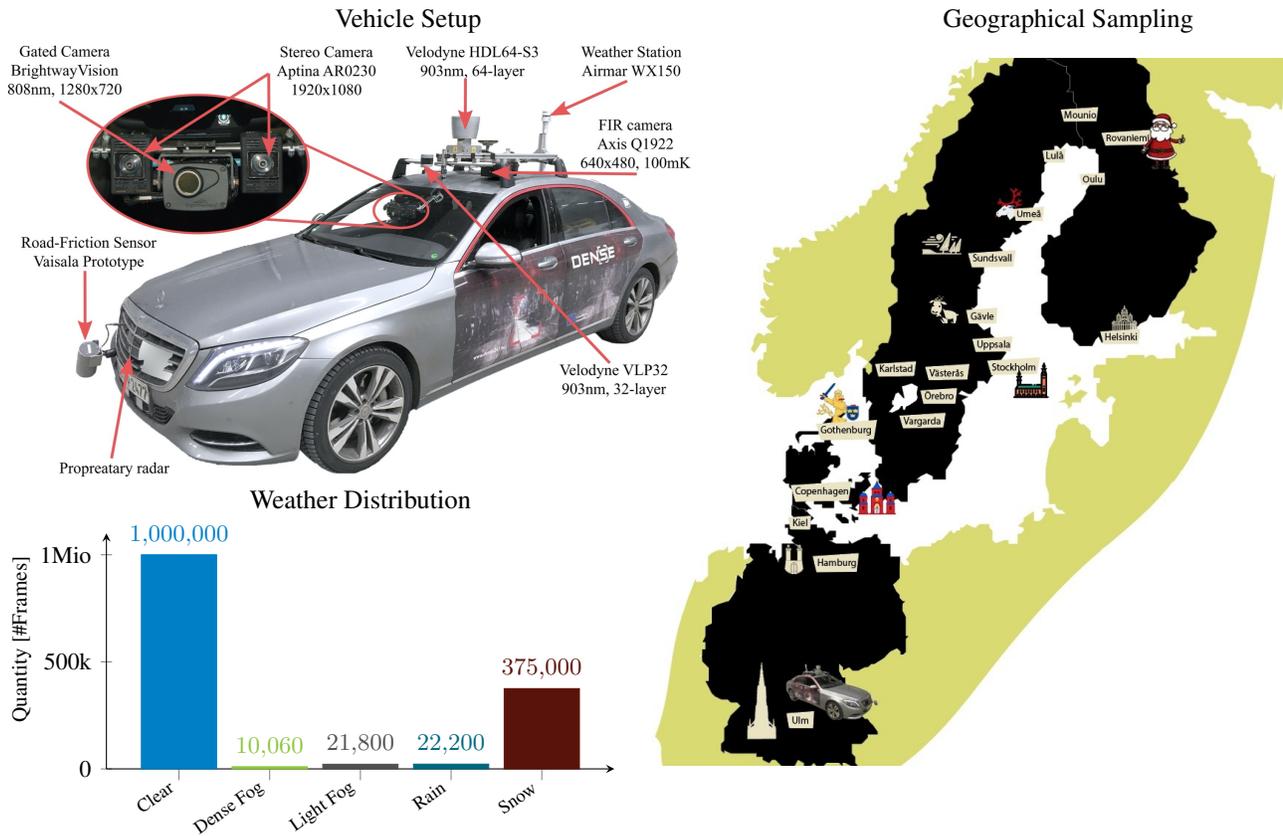
\begin{figure}[t!]
	\setlength\tabcolsep{1.5pt}
	\centering
	\includegraphics[width=\columnwidth]{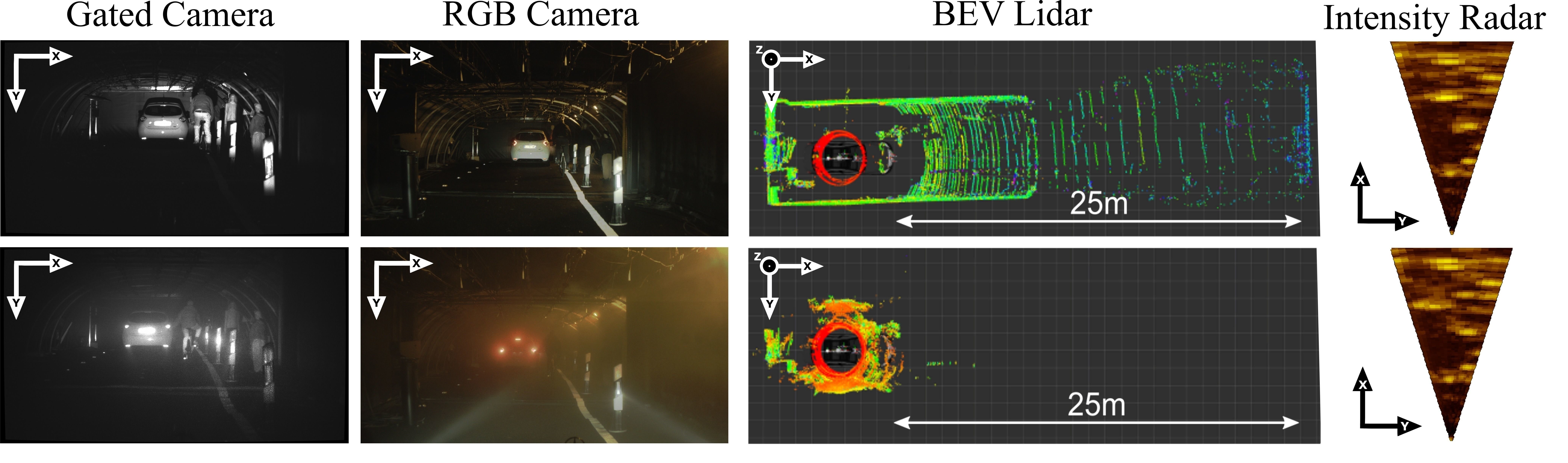}
	\vspace{-15pt} 
	\caption{Multimodal sensor response of RGB camera, scanning lidar, gated camera, and radar in a fog chamber with dense fog. Reference recordings under clear conditions are shown in the first row, recordings in fog with visibility of \unit[23]{m} are shown in the second row.}
	\label{fig:sensor_performance}
	\vspace{-2.5em}
\end{figure}
To assess object detection in adverse weather, we have acquired a large-scale automotive dataset providing  2D and 3D detection bounding boxes for multimodal data with a fine classification of weather, illumination, and scene type in rare adverse weather situations. Table~\ref{tab:SOTADatasets} compares our dataset to recent large-scale automotive datasets, such as the Waymo~\cite{waymo_open_dataset}, NuScenes~\cite{caesar2019nuscenes}, KITTI~\cite{Kitt_dataset} and the BDD~\cite{yu2018bdd100k} dataset. In contrast to \cite{caesar2019nuscenes} and \cite{yu2018bdd100k}, our dataset contains experimental data not only in light weather conditions but also in heavy snow, rain, and fog. A detailed description of the annotation procedures and label specifications is given in the supplemental material. With this cross-weather annotation of multimodal sensor data and broad geographical sampling, it is the only existing dataset that allows for the assessment of our multimodal fusion approach. In the future, we envision researchers developing and evaluating multimodal fusion methods in weather conditions not covered in existing datasets.

In Figure~\ref{fig:car}, we plot the weather distribution of the proposed dataset. The statistics were obtained by manually annotating all synchronized frames at a frame rate of \unit[0.1]{Hz}. We guided human annotators to distinguish light from dense fog when the visibility fell below \unit[1]{km}~\cite{FogDefinition} and \unit[100]{m}, respectively. If fog occurred together with precipitation, the scenes were either labeled as snowy or rainy depending on the environment road conditions. For our experiments, we combined snow and rainy conditions.
Note that the statistics validate the rarity of scenes in heavy adverse weather, which is in agreement to \cite{VanOldenborgh2010a} and demonstrates the difficulty and critical nature of obtaining such data in the assessment of truly self-driving vehicles, i.e. without the interaction of remote operators outside of geo-fenced areas. We found that extreme adverse weather conditions occur only locally and change very quickly. 

The individual weather conditions result in asymmetrical perturbations of various sensor technologies, leading to asymmetric degradation, i.e. instead of all sensor outputs being affected uniformly by a deteriorating environmental condition, some sensors degrade more than others, see Figure~\ref{fig:sensor_performance}. For example, conventional passive cameras perform well in daytime conditions, but their performance degrades in night-time conditions or challenging illumination settings such as low sun illumination. Meanwhile, active scanning sensors as lidar and radar are less affected by ambient light changes due to active illumination and a narrow bandpass on the detector side. On the other hand, active lidar sensors are highly degraded by scattering media as fog, snow or rain, limiting the maximal perceivable distance at fog densities below \unit[50]{m} to \unit[25]{m}, see Figure~\ref{fig:sensor_performance}. Millimeter-wave radar waves do not strongly scatter in fog~\cite{Hasirlioglu2016}, but currently provide only low azimuthal resolution. Recent gated images have shown robust perception in adverse weather \cite{Gated2Depth}, provide high spatial resolution, but are lacking color information compared to standard imagers. With these sensor-specific weaknesses and strengths of each sensor, multimodal data can be crucial in robust detection methods.

\subsection{Multimodal Sensor Setup}
For acquisition we have equipped a test vehicle with sensors covering the visible, mm-wave, NIR, and FIR band, see Figure~\ref{fig:car}. We measure intensity, depth, and weather condition. 

\vspace{-0.5em}
\paragraph{Stereo Camera} 
As visible-wavelength RGB cameras, we use a stereo pair of two front-facing high-dynamic-range automotive RCCB cameras, consisting of two OnSemi AR0230 imagers with a resolution of $1920\times 1024$, a baseline of \unit[20.3]{cm} and \unit[12]{bit} quantization. The cameras run at \unit[30]{Hz} and are synchronized for stereo imaging. 
Using Lensagon B5M8018C optics with a focal length of \unit[8]{mm}, a field-of-view of \unit[39.6]{\textdegree} $\times$ \unit[21.7]{\textdegree} is obtained.

\vspace{-0.5em}
\paragraph{Gated camera}
We capture gated images in the NIR band at \unit[808]{nm} using a BrightwayVision BrightEye camera operating at \unit[120]{Hz} with a resolution of $1280\times720$ and a bit depth of \unit[10]{bit}. 
The camera provides a similar field-of-view as the stereo camera with \unit[31.1]{\textdegree} $\times$ \unit[17.8]{\textdegree}.
Gated imagers rely on time-synchronized camera and flood-lit flash laser sources \cite{Inbar2008}. 
The laser pulse emits a variable narrow pulse, and the camera captures the laser echo after an adjustable delay. This enables to significantly reduce backscatter from particles in adverse weather \cite{BijelicGruberGated2018}. Furthermore, the high imager speed enables to capture multiple overlapping slices with different range-intensity profiles encoding extractable depth information in between multiple slices~\cite{Gated2Depth}. Following \cite{Gated2Depth}, we capture 3 broad slices for depth estimation and additionally 3-4 narrow slices together with their passive correspondence at a system sampling rate of \unit[10]{Hz}.

\vspace{-0.5em}
\paragraph{Radar} For radar sensing, we use a proprietary frequency-modulated continuous wave (FMCW) radar at \unit[77]{GHz} with \unit[1]{\textdegree} angular resolution and distances up to \unit[200]{m}. The radar provides position-velocity detections at \unit[15]{Hz}.

\vspace{-0.5em}
\paragraph{Lidar} On the roof of the car, we mount two laser scanners from Velodyne, namely HDL64 S3D and VLP32C. Both are operating at \unit[903]{nm} and can provide dual returns (strongest and last) at \unit[10]{Hz}. While the Velodyne HDL64 S3D provides equally distributed 64 scanning lines with an angular resolution of \unit[0.4]{\textdegree}, the Velodyne VLP32C offers 32 non-linear distributed scanning lines. HDL64 S3D and VLP32C scanners achieve a range of \unit[100]{m} and \unit[120]{m}, respectively.

\vspace{-0.5em}
\paragraph{FIR camera} Thermal images are captured with an Axis Q1922 FIR camera at \unit[30]{Hz}. The camera offers a resolution of $640\times480$ with a pixel pitch of \unit[17]{\textmu m} and a noise equivalent temperature difference (NETD) $<$ \unit[100]{mK}.

\vspace{-0.5em}
\paragraph{Environmental Sensors}
We measured environmental information with an Airmar WX150 weather station that provides temperature, wind speed and humidity, and a proprietary road friction sensor. All sensors are time-synchronized and ego-motion corrected using a proprietary inertial measurement unit (IMU). The system provides a sampling rate of \unit[10]{Hz}. 

\subsection{Recordings}

\vspace{-0.5em}
\paragraph{Real-world Recordings}
All experimental data has been captured during two test drives in February and December 2019 in Germany, Sweden, Denmark, and Finland for two weeks each, covering a distance of \unit[10,000]{km} under different weather and illumination conditions. 
A total of 1.4 million frames at a frame rate of \unit[10]{Hz} have been collected. Every 100th frame was manually labeled to balance scene type coverage. The resulting annotations contain 5,5k clear weather frames, 1k captures in dense fog, 1k captures in light fog, and 4k captures in {snow/rain}. Given the extensive capture effort, this demonstrates that training data in harsh conditions is rare. We tackle this approach by training only on clear data and testing on adverse data. The train and test regions do not have any geographic overlap. Instead of partitioning by frame, we partition our dataset based on independent recordings (5-60~min in length) from different locations. These recordings originate from 18 different major cities illustrated in Figure~\ref{fig:car} and several smaller cities along the route.

\vspace{-0.5em}
\paragraph{Controlled Condition Recordings}\label{Sec:fogchambermeasurements}
To collect image and range data under controlled conditions, we also provide measurements acquired in a fog chamber. Details on the fog chamber setup can be found in~\cite{Duthon2016, FogChamberClermont}. We have captured 35k frames at a frame rate of \unit[10]{Hz} and labeled a subset of 1,5k frames under two different illumination conditions (day/night) and three fog densities with meteorological visibilities $V$ of \unit[30]{m}, \unit[40]{m} and \unit[50]{m}. Details are given in the supplemental material, where we also do comparisons to a simulated dataset, using the forward model from \cite{sakaridis2018semantic}.

%% file: tikz/dataset_sweden_distribution.tikz
\begin{tikzpicture}[font=\small]
	\begin{axis}[
		xmode=linear,
		ymode=linear,
	  scaled ticks=false, 
	  tick label style={/pgf/number format/fixed},
		width=1\columnwidth,
		height=0.27\textwidth,
		bar width = 1cm,
		axis x line=bottom,axis y line=left,
		ybar=-1cm,
		title={\normalsize{{Weather Distribution}}},
		title style = {align=center},
		enlarge y limits={upper, value=0.1},
		enlarge x limits = 0.2,
		ymajorgrids = false,
		xmajorgrids = false, 
		ymin=0,
		ymax=1000000,
		nodes near coords,
		nodes near coords align={anchor=south},
		every node near coord/.append style={anchor=center, font=\tiny},
		every node near coord/.style={/pgf/number format/fixed},
		x tick label style={rotate=35, anchor=east, align=center, font=\small},
		xtick={1,2,3,4,5},
		xticklabels={\scriptsize{Clear},\scriptsize{Dense Fog},\scriptsize{Light Fog},\scriptsize{Rain},\scriptsize{Snow}},
		yticklabels={0, 0, 500k, 1Mio}
		]
		\addplot[color=mittelblau, line width=0.01cm, fill=mittelblau] coordinates {
			(1, 1000000)}; 
		\addplot[color=apfelgruen, line width=0.01cm, fill=apfelgruen] coordinates {
			(2, 10060)};
		\addplot[color=dunkelgrau, line width=0.01cm, fill=dunkelgrau] coordinates {
			(3, 21800)}; 
		\addplot[color=dai_petrol, line width=0.01cm, fill=dai_petrol] coordinates {
			(4, 22200)}; 
		\addplot[color=dai_deepred20K, line width=0.01cm, fill=dai_deepred20K] coordinates {
			(5, 375000)}; 
		\end{axis}
\end{tikzpicture}

%% file: method.tex
\begin{figure*}[ht!]
    \centering
        \includegraphics[width=0.98\textwidth]{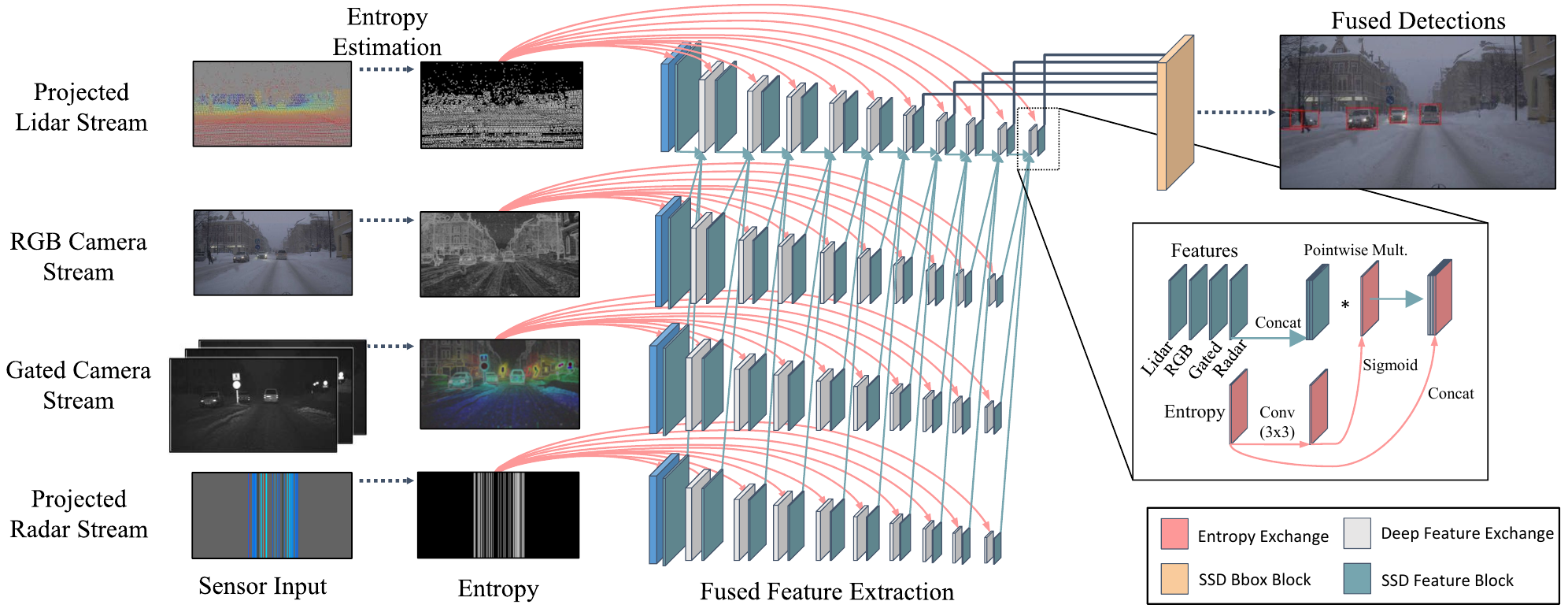}
	    \vspace{-0.8em}   
    \caption{Overview of our architecture consisting of four single-shot detector branches with deep feature exchange and adaptive fusion of lidar, RGB camera, gated camera, and radar. All sensory data is projected into the camera coordinate system following Sec.~\ref{sec:datarep}. To steer fusion in-between sensors, the model relies on sensor entropy, which is provided to each feature exchange block \textit{(red)}. The deep feature exchange blocks \textit{(white)}  interchange information \textit{(blue)} with parallel feature extraction blocks. The fused feature maps are analyzed by SSD blocks \textit{(orange)}.}\label{fig:arch}
     \vspace{-1.5em}   
\end{figure*}
In this section, we describe the proposed adaptive deep fusion architecture that allows for multimodal fusion in the presence of unseen asymmetric sensor distortions. We devise our architecture under real-time processing constraints required for self-driving vehicles and autonomous drones. Specifically, we propose an efficient single-shot fusion architecture.
\subsection{Adaptive Multimodal Single-Shot Fusion}
The proposed network architecture is shown in Figure~\ref{fig:arch}. It consists of multiple single-shot detection branches, each analyzing one sensor modality.

\vspace{-0.5em}
\paragraph{Data Representation}\label{sec:datarep} The camera branch uses conventional three-plane RGB inputs, while for the lidar and radar branch, we depart from recent bird's eye-view (BeV) projection~\cite{ku2018joint} schemes or raw point-cloud representations~\cite{xu2017pointfusion}. BeV projection or point-cloud inputs do not allow for deep early fusion as the feature representations in the early layers are inherently different from the camera features. Hence, existing BeV fusion methods can only fuse features in a lifted space, after matching region proposals, but not earlier. Figure~\ref{fig:arch} visualizes the proposed input data encoding, which aids deep multimodal fusion. 
Instead of using a naive depth-only input encoding, we provide depth, height, and pulse intensity as input to the lidar network. 
For the radar network, we assume that the radar is scanning in a 2D-plane orthogonal to the image plane and parallel to the horizontal image dimension. Hence, we consider radar invariant along the vertical image axis and replicate the scan along vertical axis. Gated images are transformed into the image plane of the RGB camera using a homography mapping, see supplemental material. The proposed input encoding allows for a position and intensity-dependent fusion with pixel-wise correspondences between different streams. We encode missing measurement samples with zero value.

\vspace{-0.5em}
\paragraph{Feature Extraction} As feature extraction stack in each stream, we use a modified VGG~\cite{simonyan2014very} backbone. Similar to \cite{ku2018joint,chen2017multi}, we reduce the number of channels by half and cut the network at the conv4 layer. Inspired by \cite{SSDLiu2015,featurePyramid}, we use six feature layers from conv4-10 as input to SSD detection layers. The feature maps decrease in size\footnote{\tiny{We use a feature map pyramid $\left[(24, 78), (24, 78), (12, 39), (12, 39), (6, 20), (3, 10)\right]$}}, implementing a feature pyramid for detections at different scales. As shown in Figure~\ref{fig:arch}, the activations of different feature extraction stacks are exchanged. To steer fusion towards the most reliable information, we provide the sensor entropy to each feature exchange block. We first convolve the entropy, apply a sigmoid, multiply with the concatenated input features from all sensors, and finally concatenate the input entropy. The folding of entropy and application of the sigmoid generates a multiplication matrix in the interval [0,1]. This scales the concatenated features for each sensor individually based on the available information. Regions with low entropy can be attenuated, while entropy rich regions can be amplified in the feature extraction. Doing so allows us to \emph{adaptively fuse features} in the feature extraction stack itself, which we motivate in depth in the next section. 

\begin{figure*}[t]
	\hspace{2pt}\begin{tikzpicture}
	\node[black, anchor=west] at (120pt,135 pt) {{\includegraphics[trim=0 0 0 0, clip, width=0.25\columnwidth]{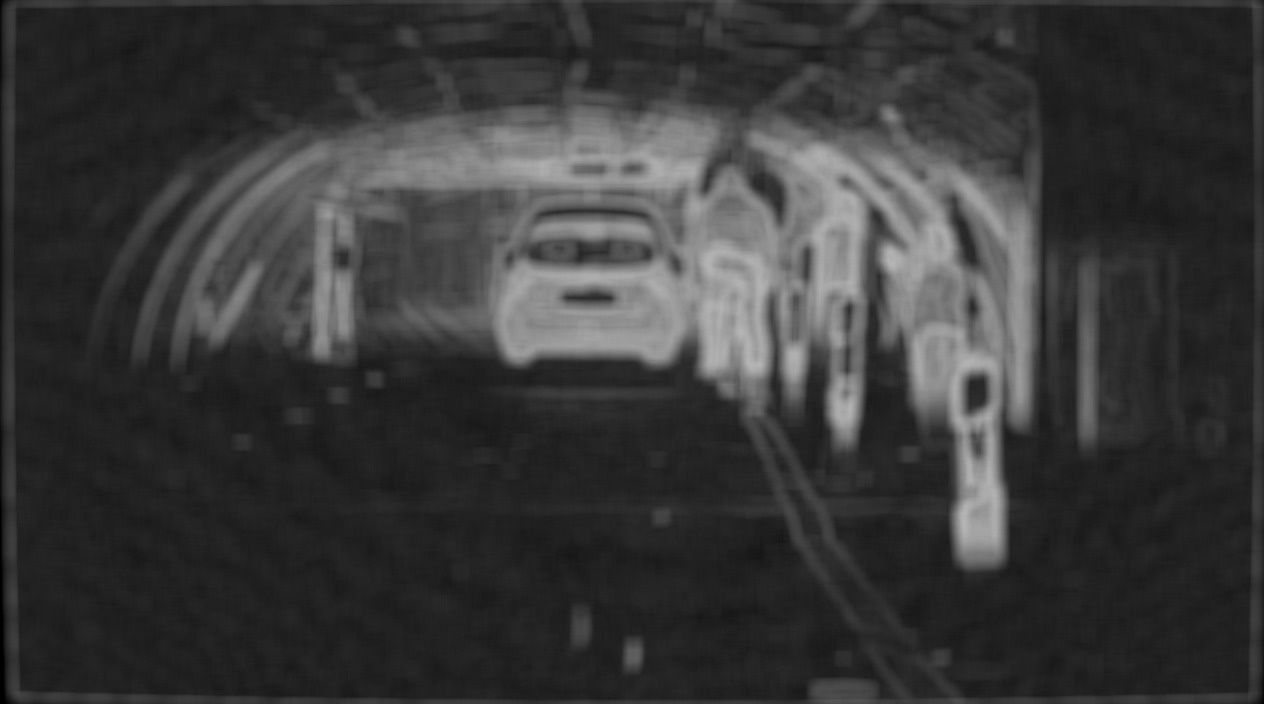}}};
	\node[mark size=3pt,color=dai_ligth_grey40K] at (130pt,145 pt) {\pgfuseplotmark{*}};
	\node[black, anchor=west] at (120pt,100 pt) {{\includegraphics[trim=0 0 0 0, clip, width=0.25\columnwidth]{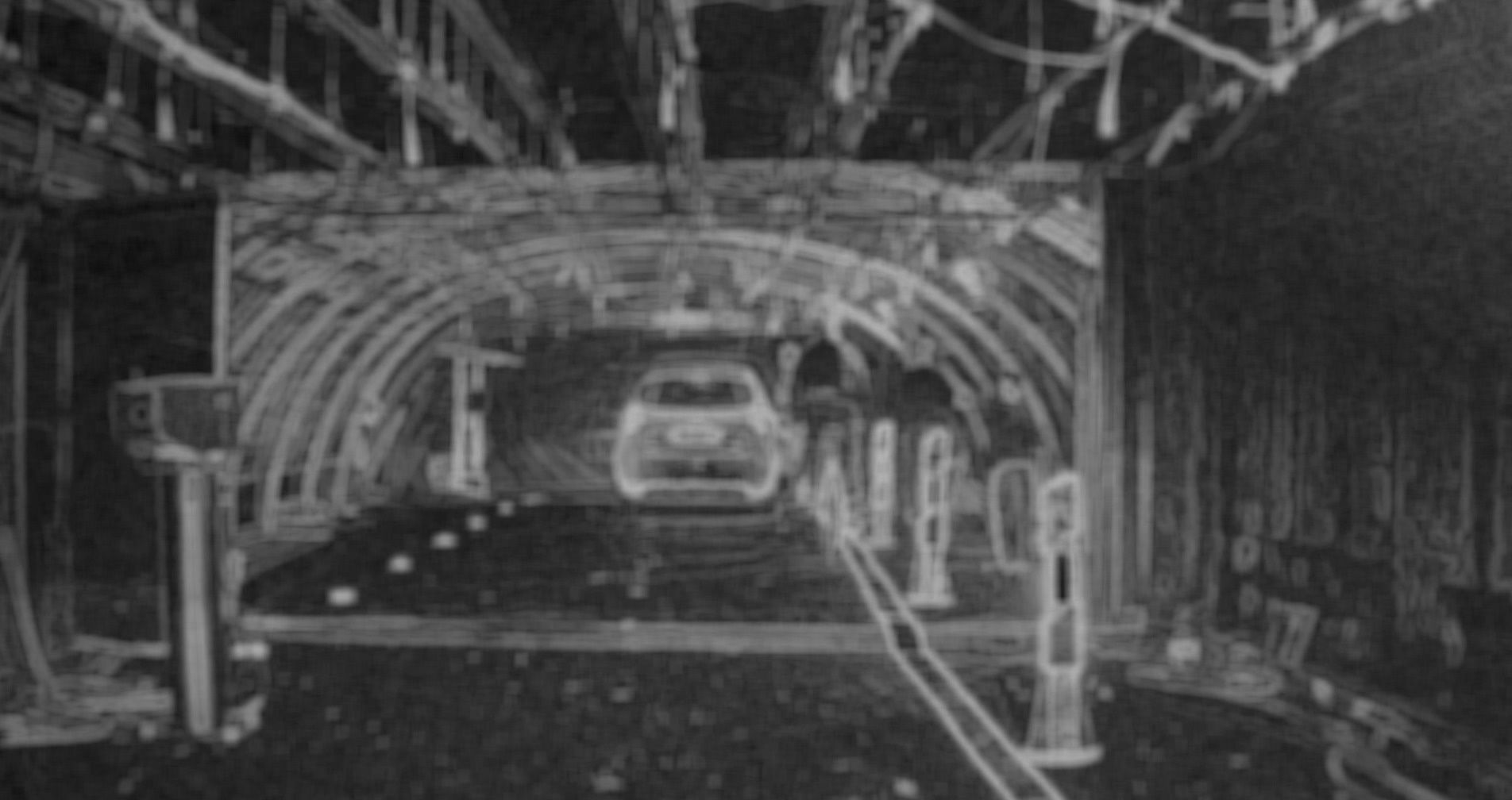}}};
	\node[mark size=3pt,color=dai_deepred] at (127pt,110 pt) {\pgfuseplotmark{triangle*}};
	\node[black, anchor=west] at (120pt,65 pt) {{\includegraphics[trim=0 0 0 0, clip, width=0.25\columnwidth]{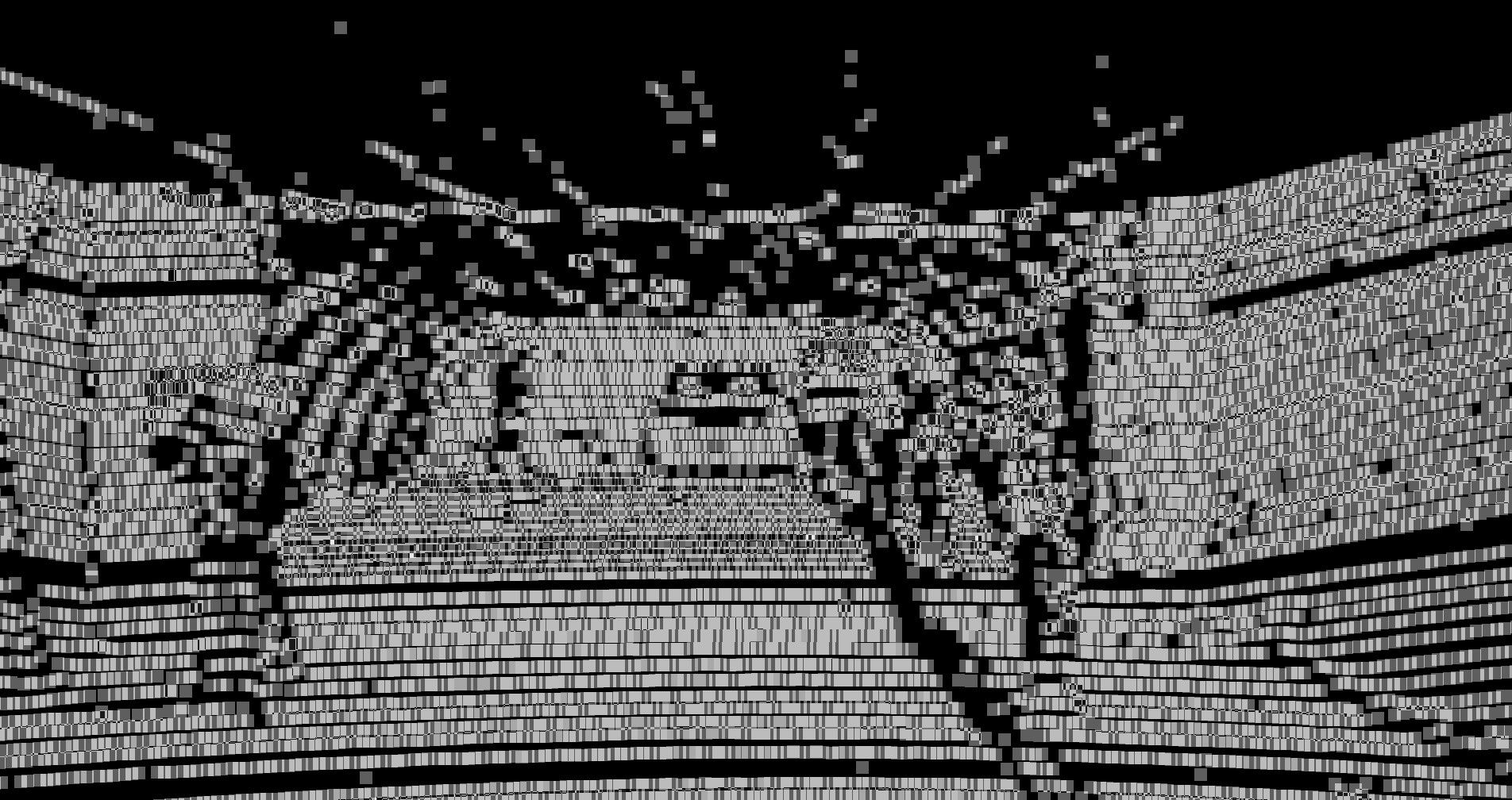}}};
	\node[mark size=3pt,color=dai_petrol] at (130pt,75 pt) {\pgfuseplotmark{square*}};
	\node[black, anchor=west] at (120pt,30 pt) {{\includegraphics[trim=0 0 0 0, clip, width=0.25\columnwidth]{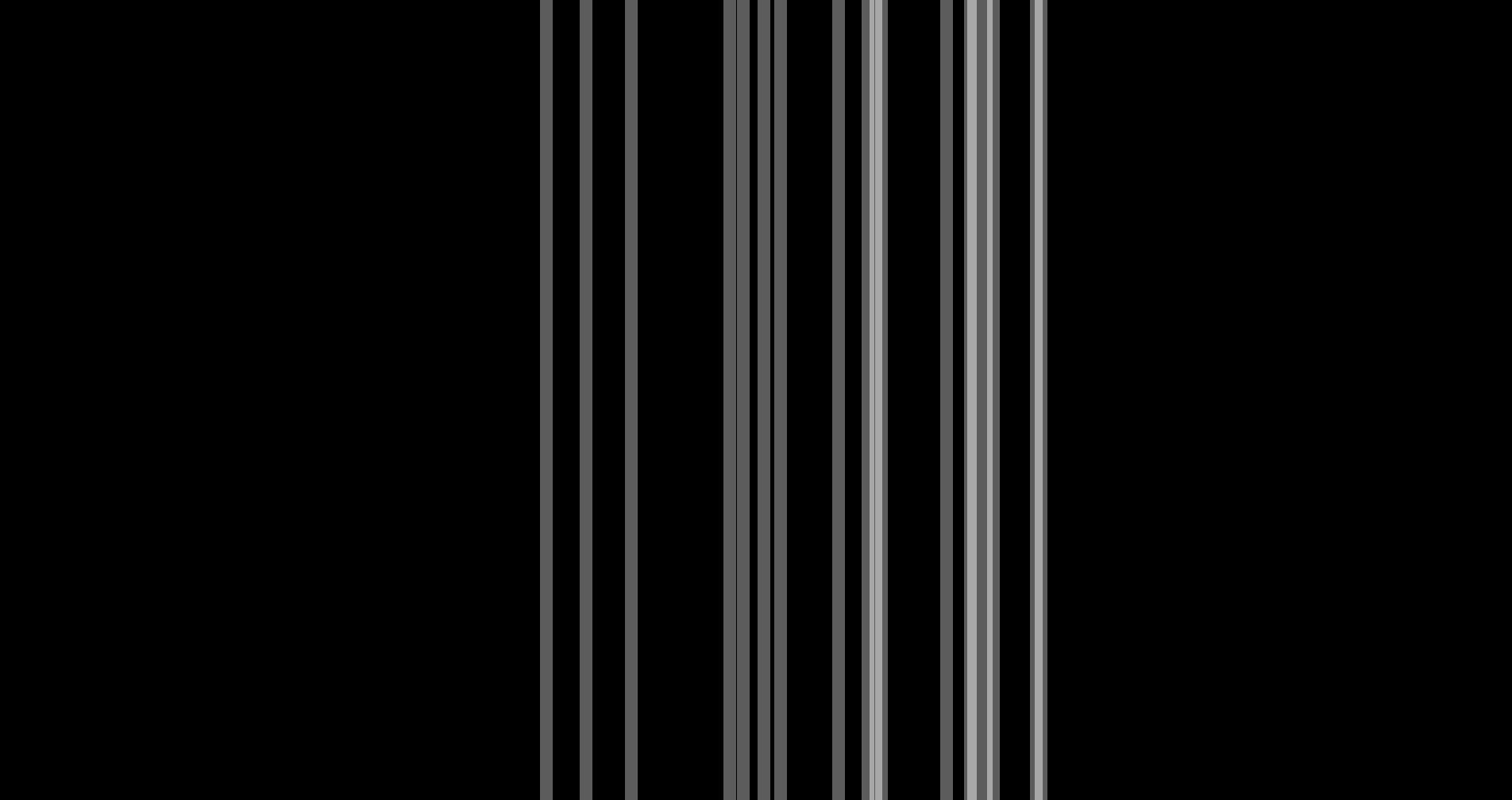}}};
	\node[mark size=3pt,color=magenta] at (130pt,40 pt) {\pgfuseplotmark{square*}};
	\node[black, anchor=west] at (185pt,135 pt) {{\includegraphics[trim=0 0 0 0, clip, width=0.25\columnwidth]{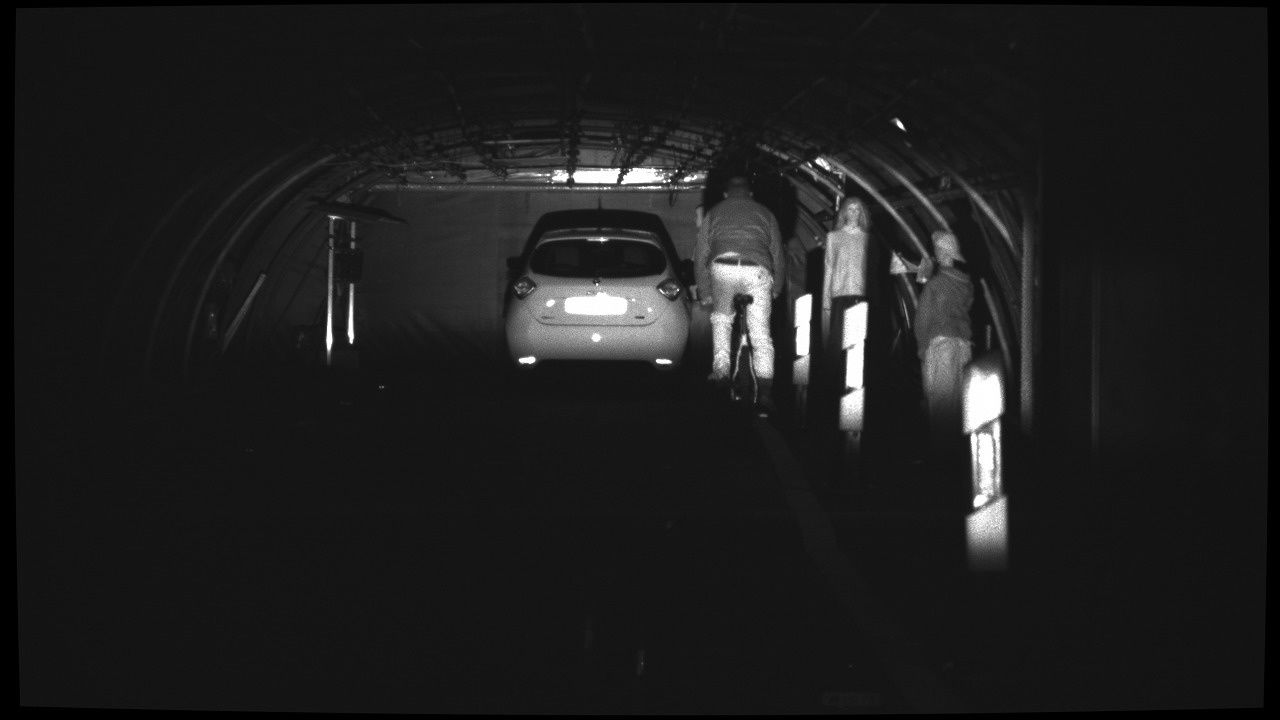}}};
	\node[very thick, mark size=3pt,color=dai_ligth_grey40K] at (195pt,145 pt) {\pgfuseplotmark{o}};
	\node[black, anchor=west] at (185pt,100 pt) {{\includegraphics[trim=0 0 0 0, clip, width=0.25\columnwidth]{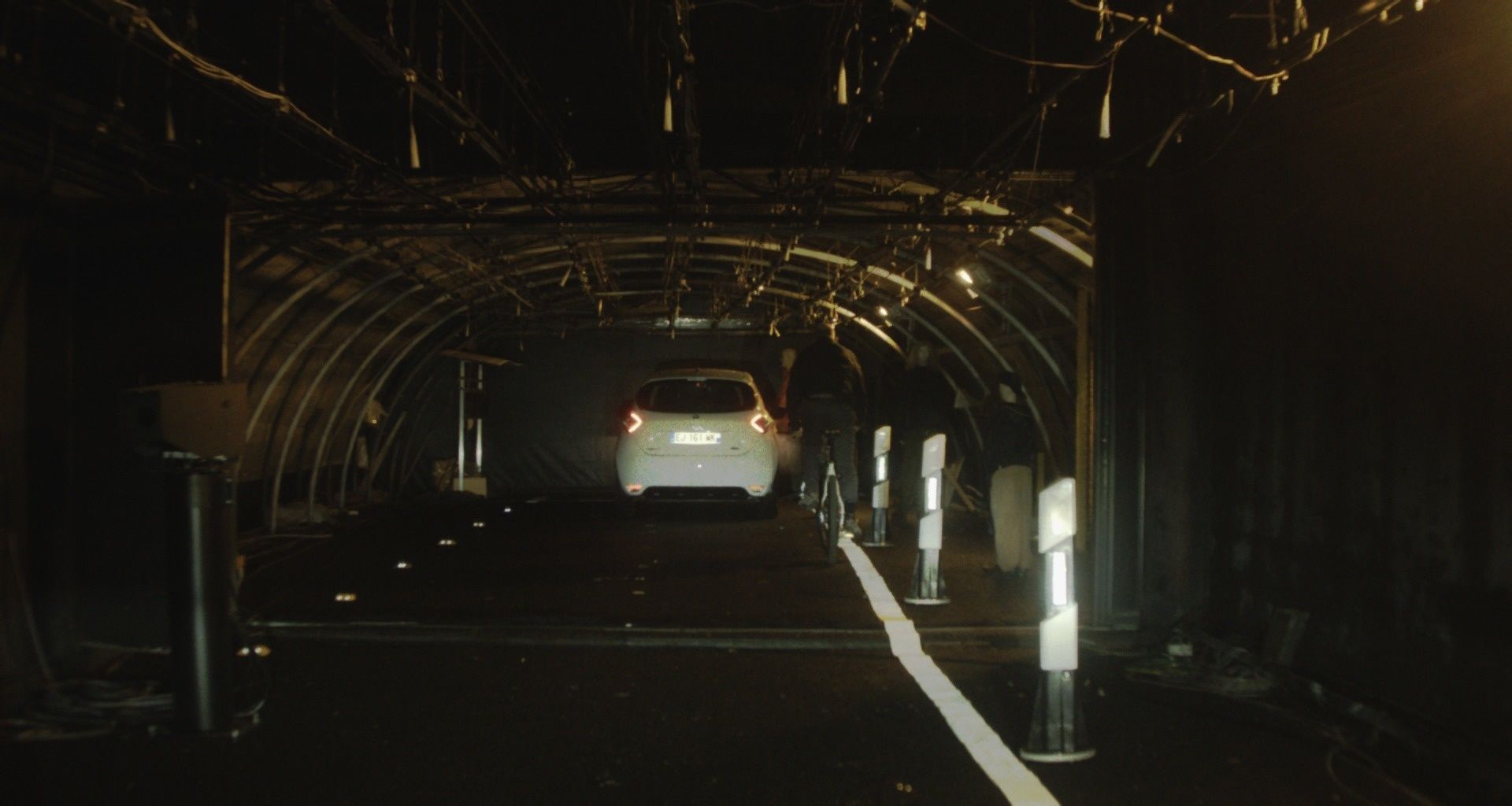}}};
	\node[very thick, mark size=3pt,color=dai_deepred] at (192pt,110 pt) {\pgfuseplotmark{triangle}};
	\node[black, anchor=west] at (185pt,65 pt) {{\includegraphics[trim=0 0 0 0, clip, width=0.25\columnwidth]{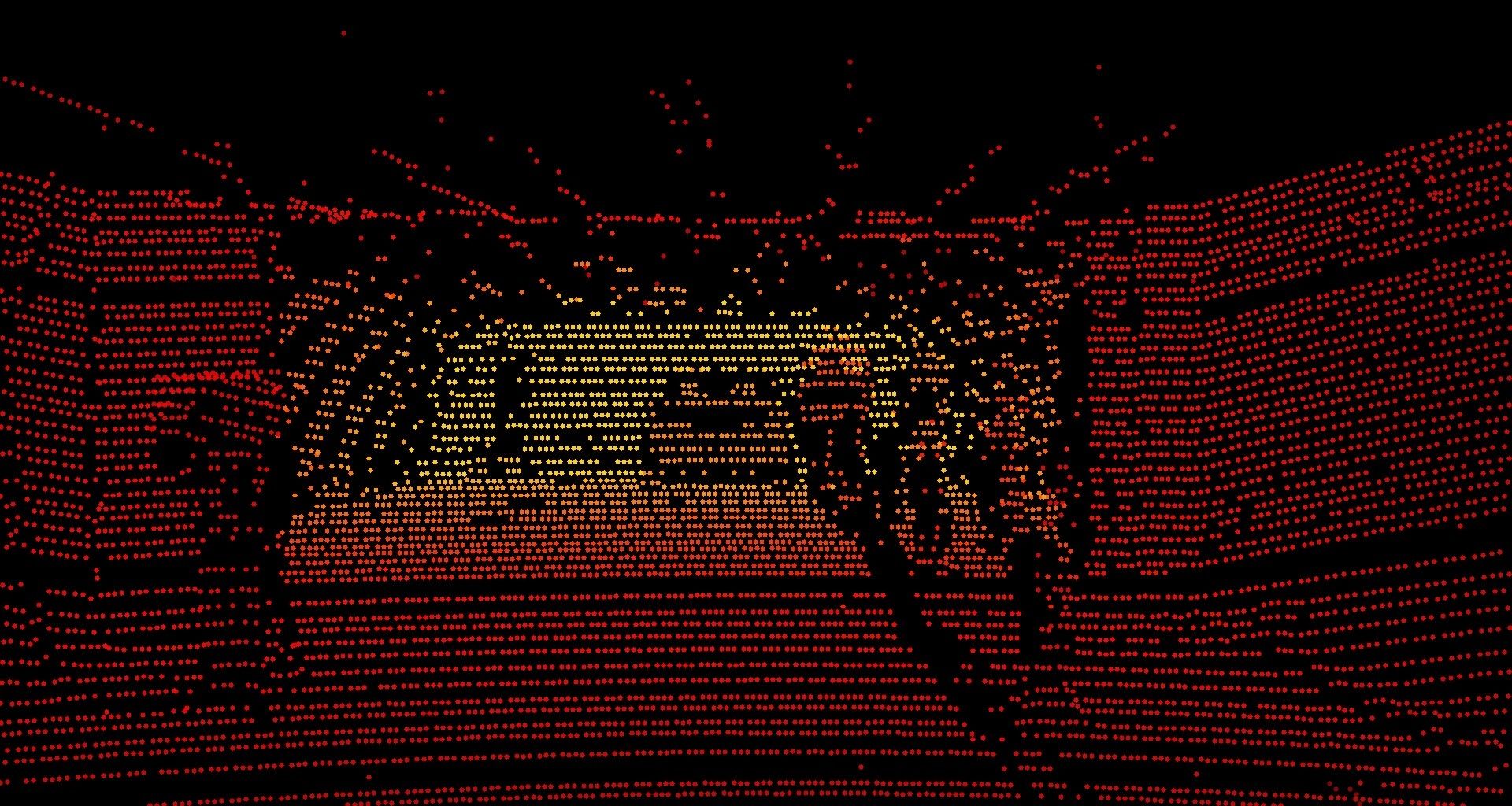}}};
	\node[very thick, mark size=3pt,color=dai_petrol] at (195pt,75 pt) {\pgfuseplotmark{square}};
	\node[black, anchor=west] at (185pt,30 pt) {{\includegraphics[trim=0 0 0 0, clip, width=0.25\columnwidth]{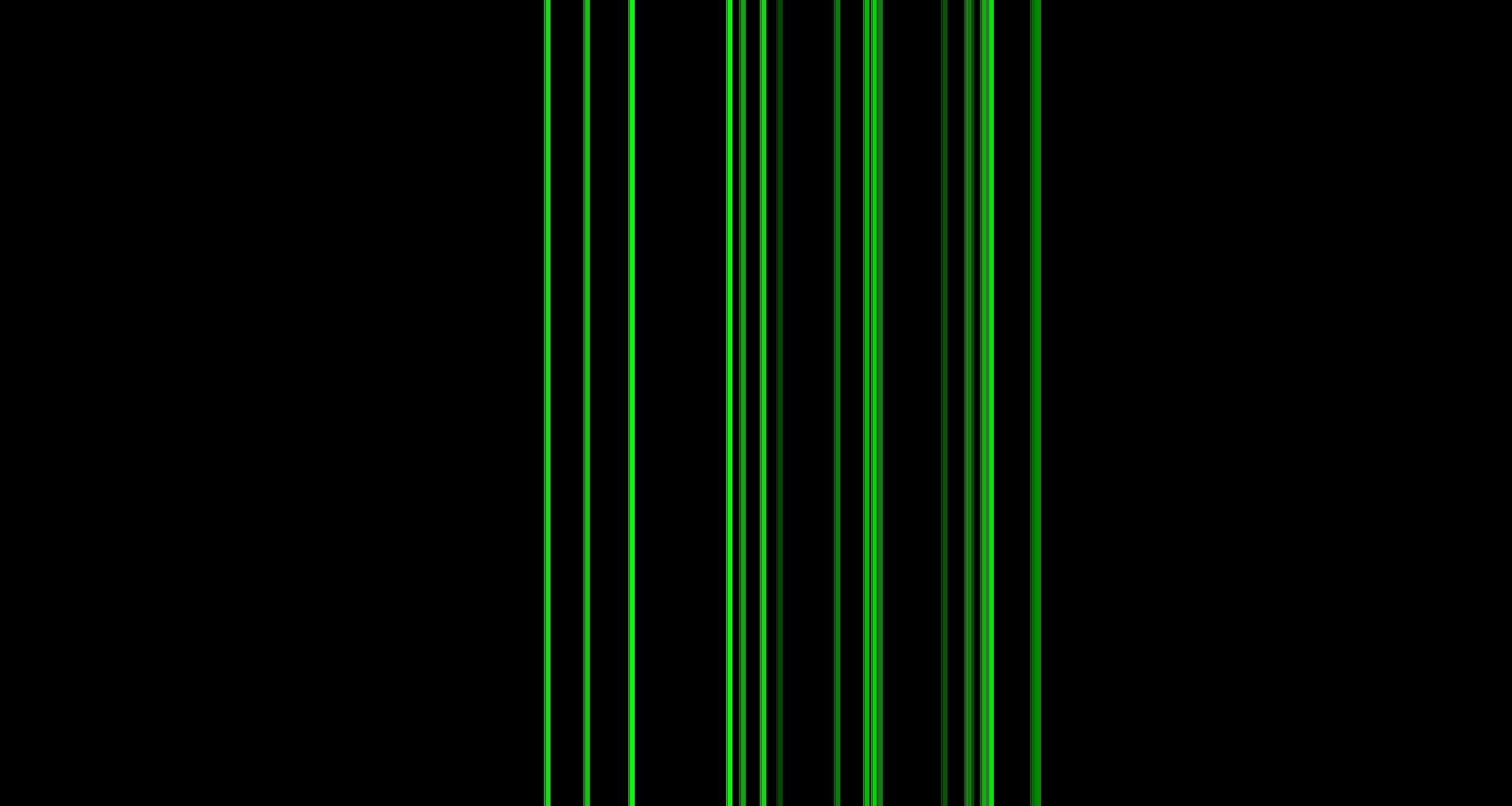}}};
	\node[very thick, mark size=3pt,color=magenta] at (195pt,40 pt) {\pgfuseplotmark{square}};
	\draw [
	    thick,
	    decoration={
	        brace,
	    },
	    decorate
	] (122pt, 155pt) -- (250pt, 155pt);
	\draw[line width=0.5pt, densely dotted, -to]    (186pt,157pt) .. controls (186pt,170pt) and (215pt,157pt)  .. (215pt,184pt);
	\node[black, anchor=west] at (-10pt,135 pt) {{\includegraphics[trim=0 0 0 0, clip, width=0.25\columnwidth]{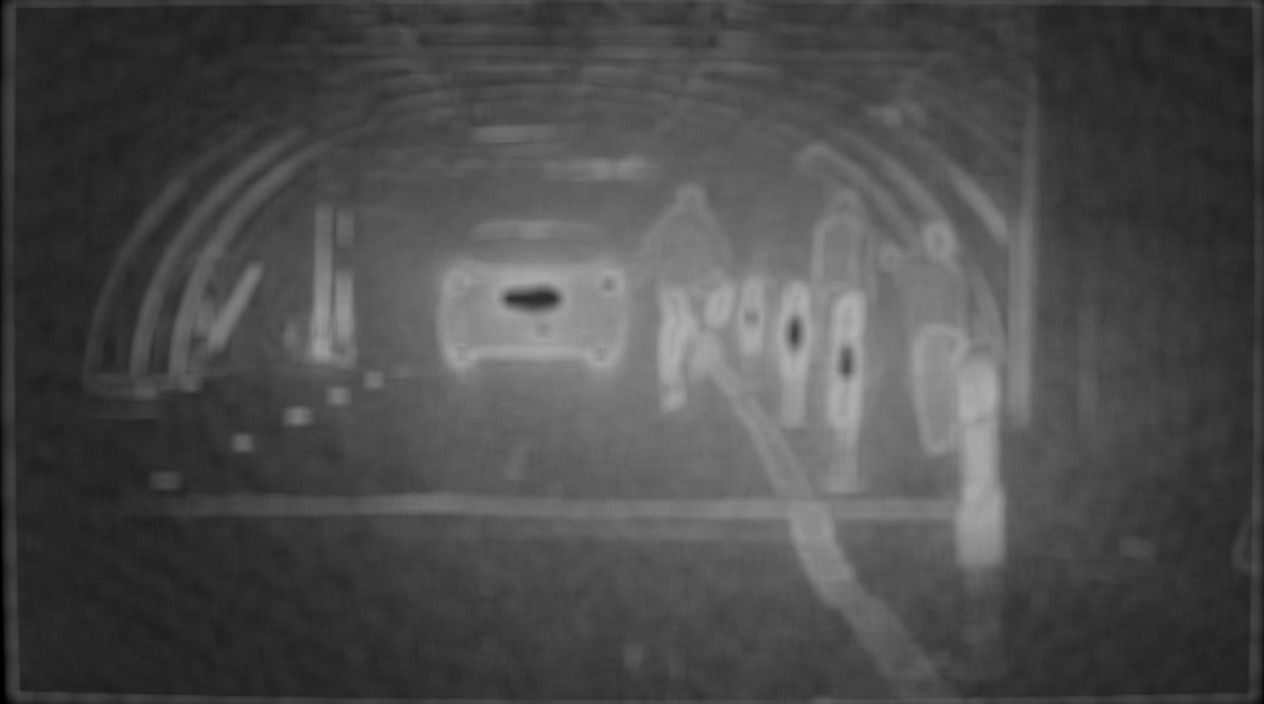}}};
	\node[black, anchor=west] at (-10pt,100 pt) {{\includegraphics[trim=0 0 0 0, clip, width=0.25\columnwidth]{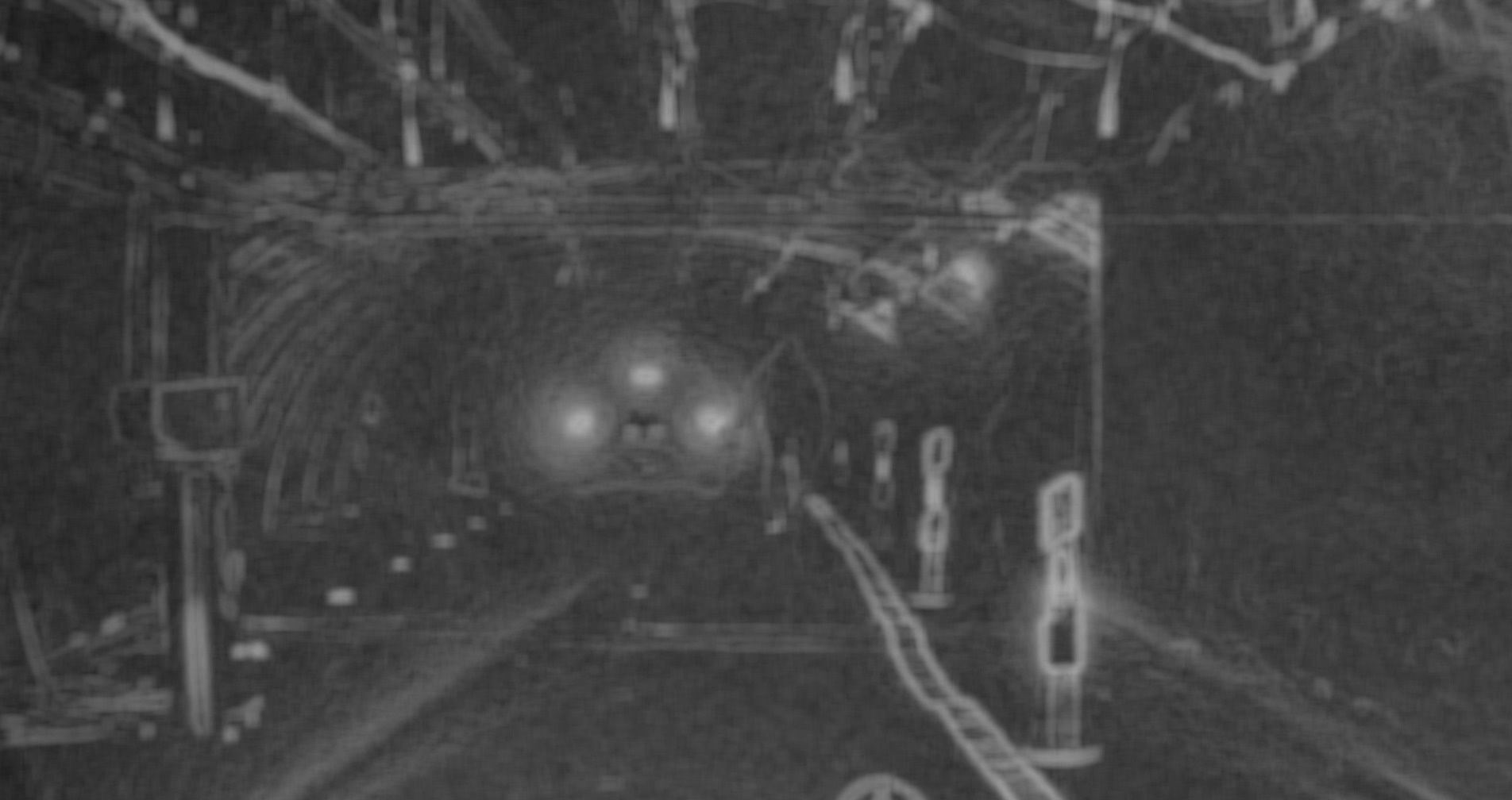}}};
	\node[black, anchor=west] at (-10pt,65 pt) {{\includegraphics[trim=0 0 0 0, clip, width=0.25\columnwidth]{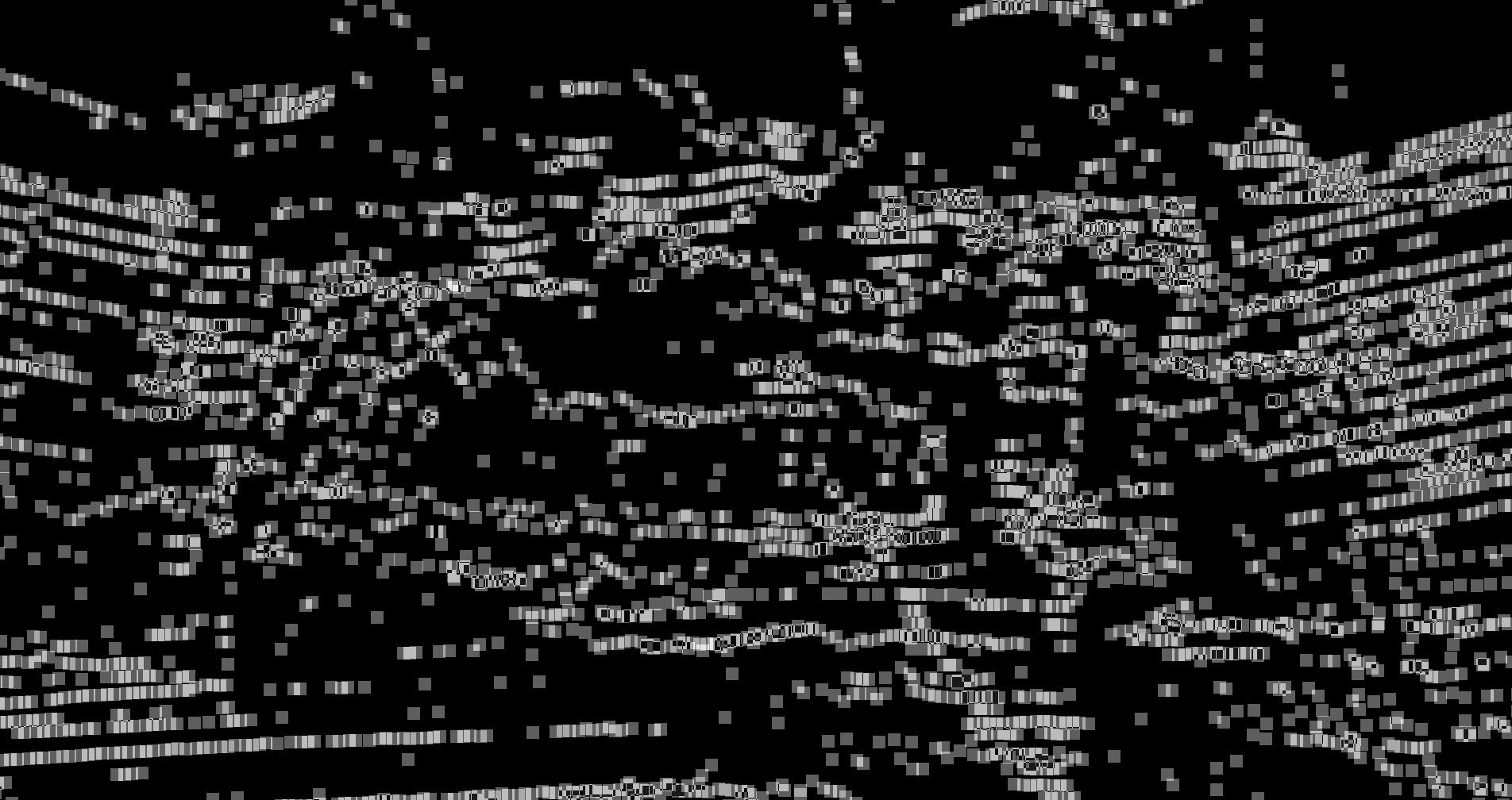}}};
	\node[black, anchor=west] at (-10pt,30 pt) {{\includegraphics[trim=0 0 0 0, clip, width=0.25\columnwidth]{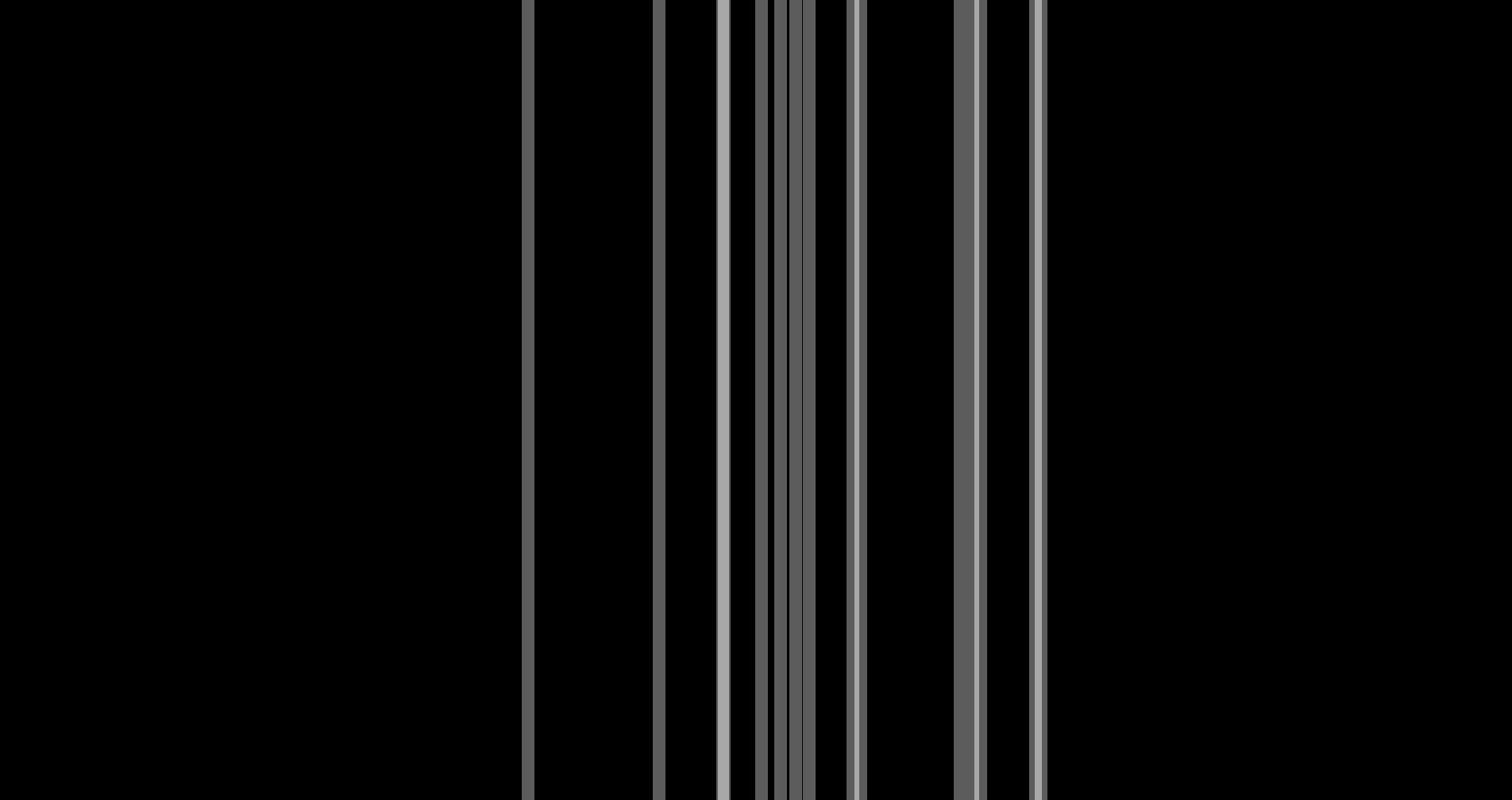}}};
	\node[black, anchor=west] at (55pt,135 pt) {{\includegraphics[trim=0 0 0 0, clip, width=0.25\columnwidth]{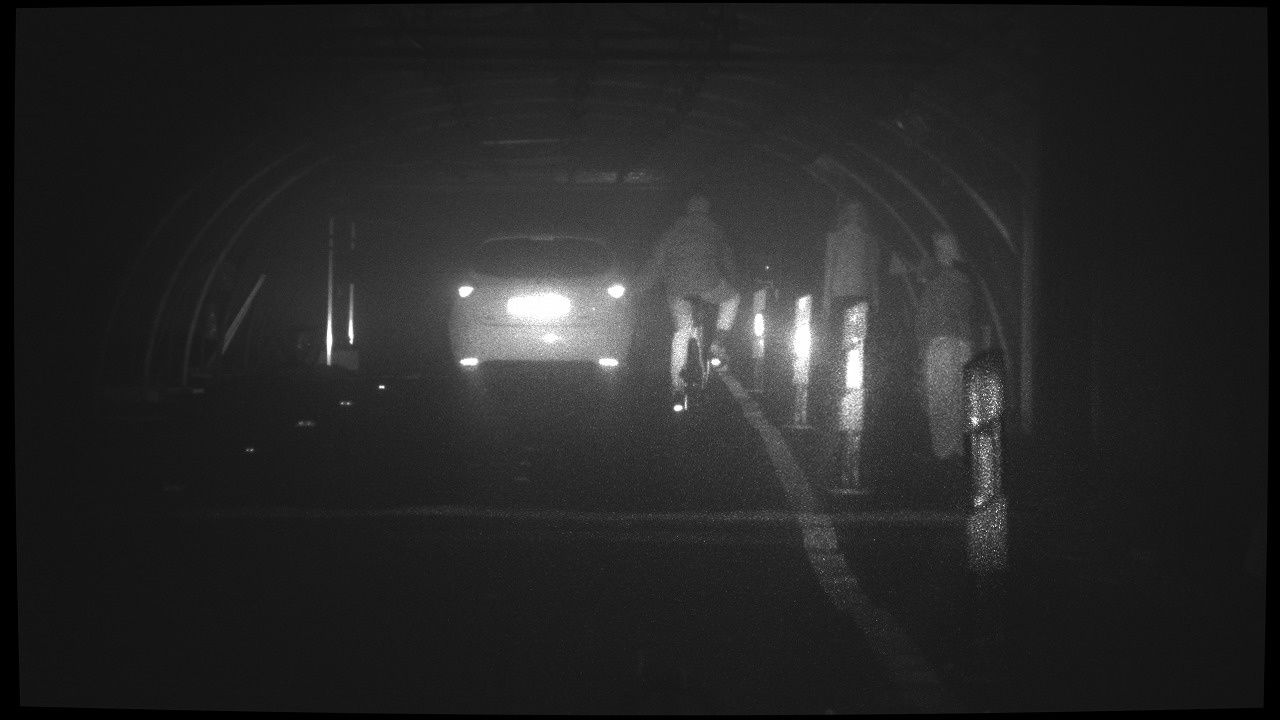}}};
	\node[black, anchor=west] at (55pt,100 pt) {{\includegraphics[trim=0 0 0 0, clip, width=0.25\columnwidth]{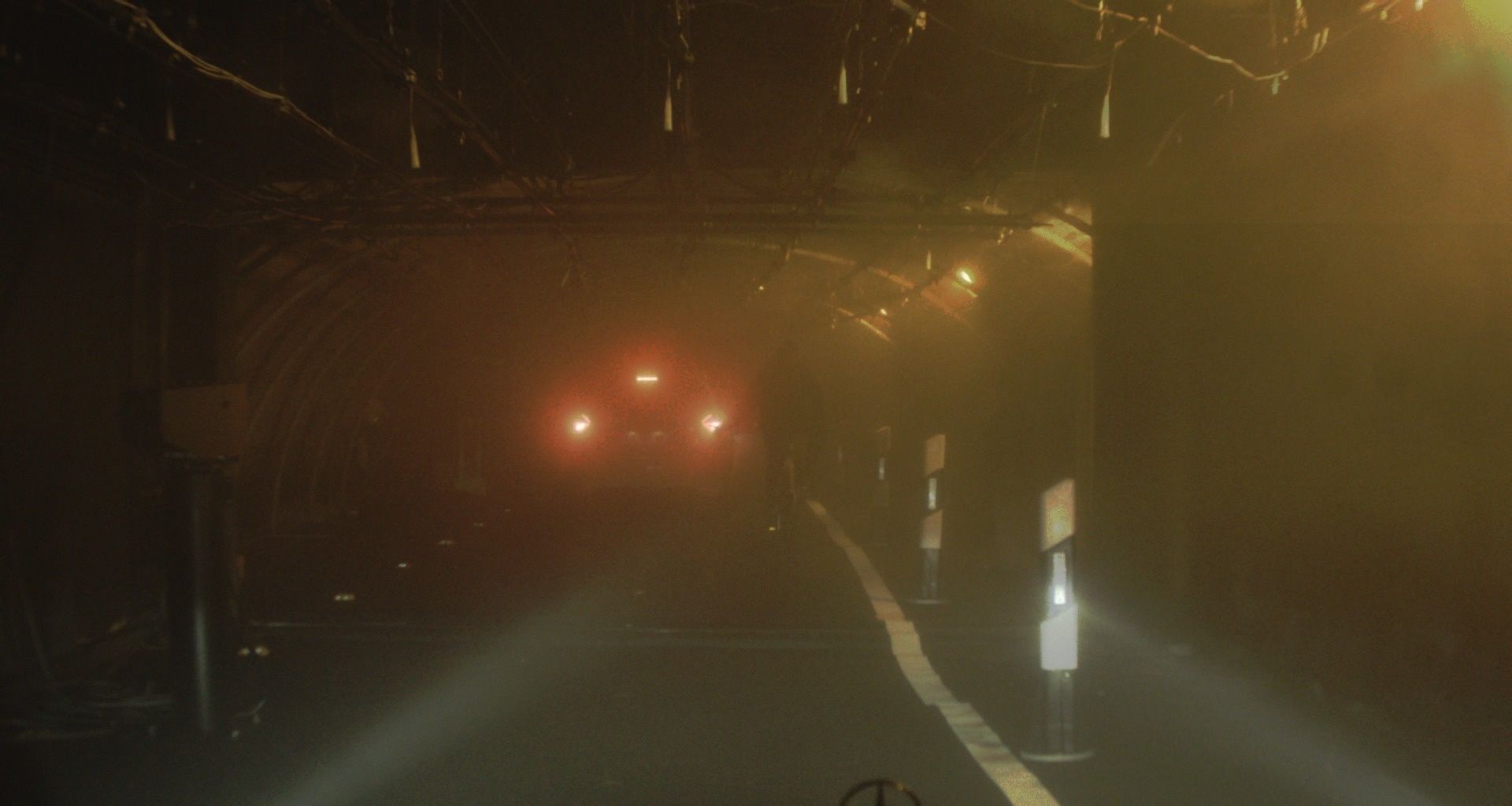}}};
	\node[black, anchor=west] at (55pt,65 pt) {{\includegraphics[trim=0 0 0 0, clip, width=0.25\columnwidth]{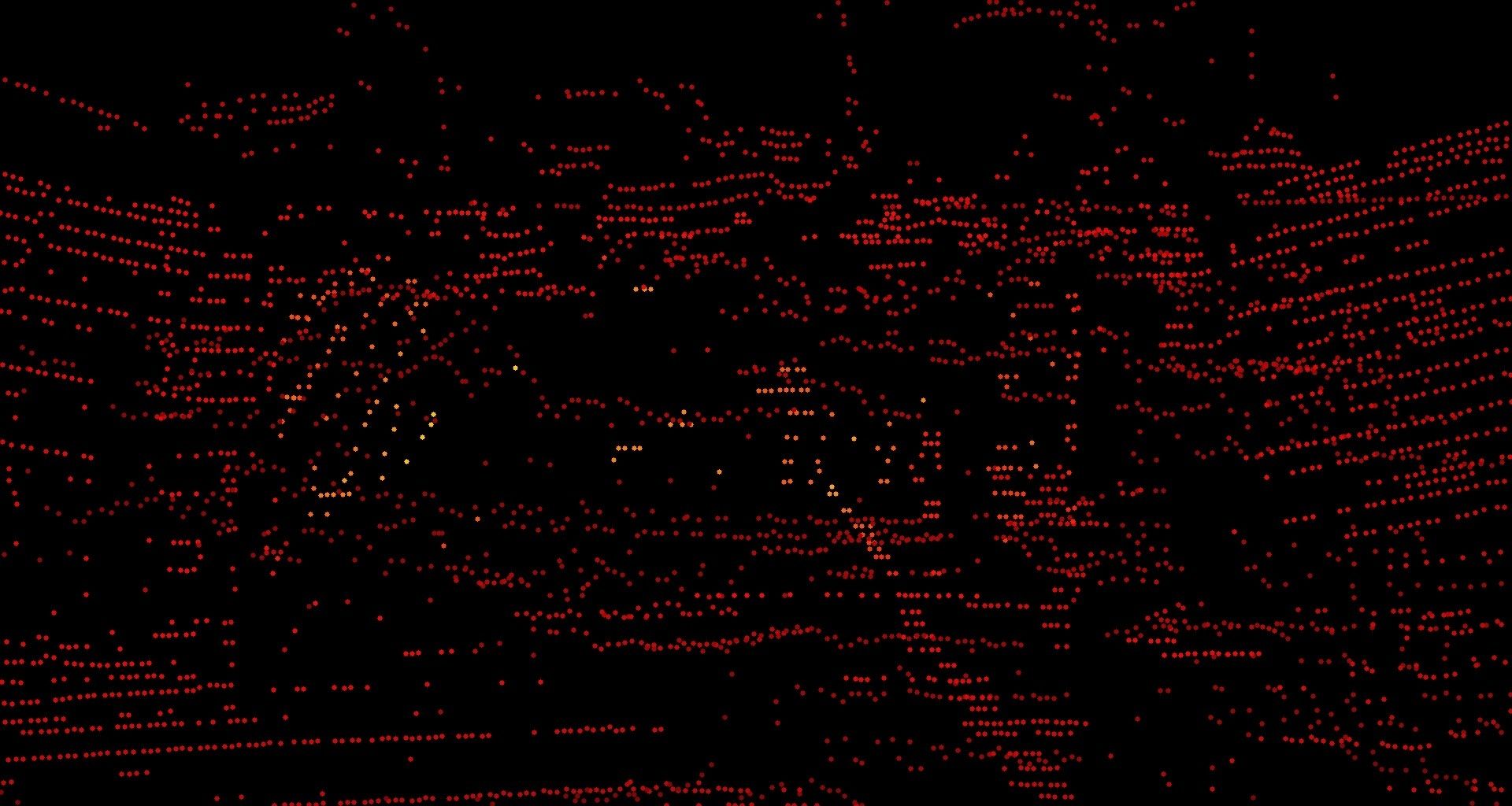}}};
	\node[black, anchor=west] at (55pt,30 pt) {{\includegraphics[trim=0 0 0 0, clip, width=0.25\columnwidth]{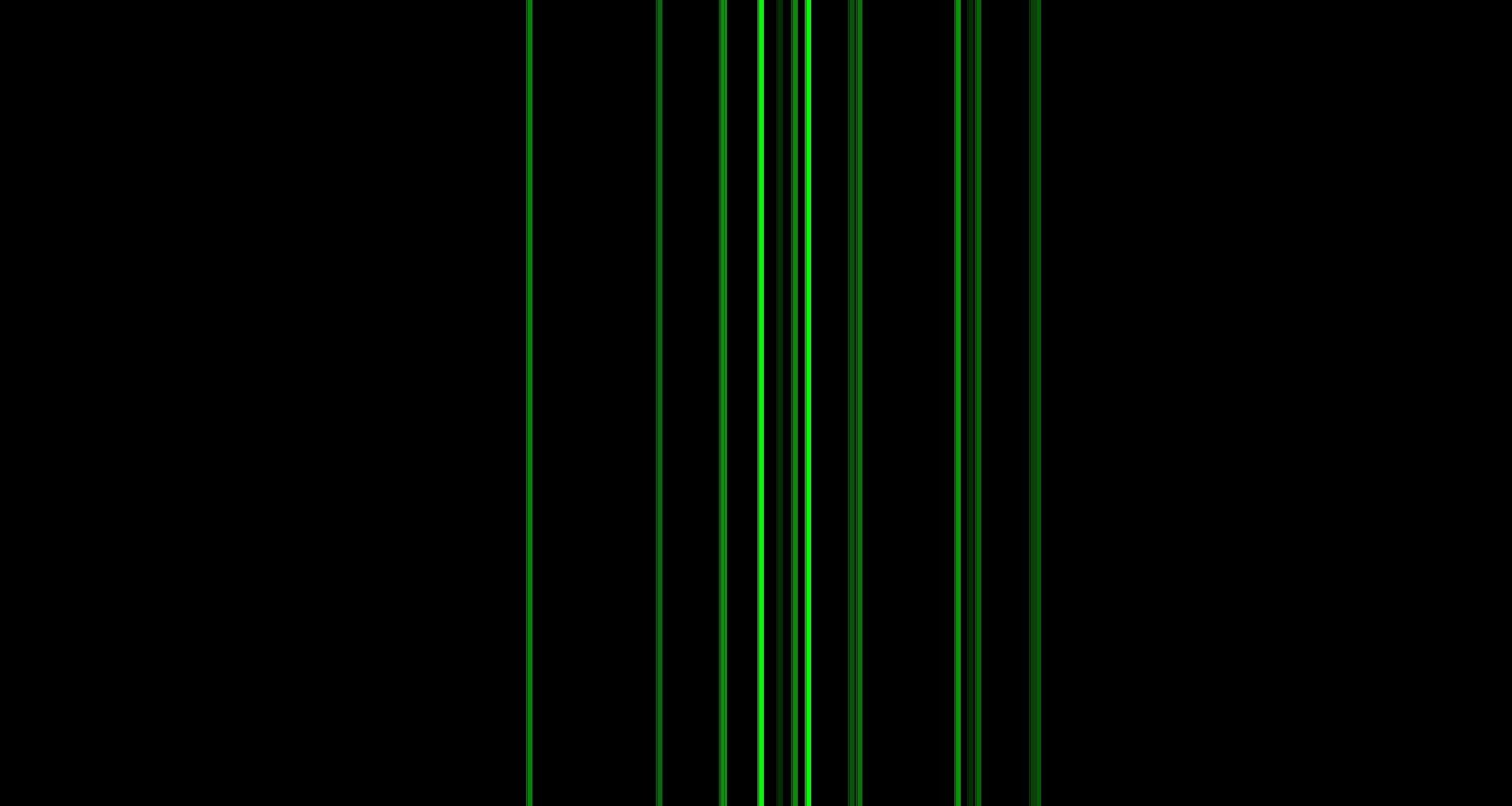}}};
	\node[very thick, mark size=3pt,color=dai_ligth_grey40K] at (65pt,145 pt) {\pgfuseplotmark{o}};
	\node[very thick, mark size=3pt,color=dai_deepred] at (62pt,110 pt) {\pgfuseplotmark{triangle}};
	\node[very thick, mark size=3pt,color=dai_petrol] at (65pt,75 pt) {\pgfuseplotmark{square}};
	\node[very thick, mark size=3pt,color=magenta] at (65pt,40 pt) {\pgfuseplotmark{square}};
	\node[mark size=3pt,color=dai_ligth_grey40K] at (0pt,145 pt) {\pgfuseplotmark{*}};
	\node[mark size=3pt,color=dai_deepred] at (-3pt,110 pt) {\pgfuseplotmark{triangle*}};
	\node[mark size=3pt,color=dai_petrol] at (0pt,75 pt) {\pgfuseplotmark{square*}};
	\node[mark size=3pt,color=magenta] at (0pt,40 pt) {\pgfuseplotmark{square*}};
	\draw [
	    thick,
	    decoration={
	        brace,
	    },
	    decorate
	] (-8pt, 155pt) -- (120pt, 155pt);
	\draw[line width=0.5pt, densely dotted, -to]  (56pt,157pt) .. controls (56pt,170pt) and (23pt,157pt)  .. (23pt,184pt);
		\node[anchor=west] (plot1) at (60pt,-5pt) {\begin{tikzpicture}
		    \begin{customlegend}[legend columns=5,legend style={draw=none,column sep=1ex},
	    	legend entries={Projected Entropies:, Gated, Image, Lidar, Radar, Projected Measurements:, Gated, Image, Lidar, Radar}]
		\addlegendimage{very thick, densely dashed, white}
		\addlegendimage{very thick, dai_ligth_grey40K, mark=*}
		\addlegendimage{very thick, dai_deepred, mark=triangle*}
		\addlegendimage{very thick, dai_petrol, mark=square*}
		\addlegendimage{very thick, magenta, mark=square*}
		\addlegendimage{very thick, densely dashed, white}
		\addlegendimage{only marks,very thick,dai_ligth_grey40K, mark=o}
		\addlegendimage{only marks,very thick,dai_deepred, mark=triangle}
		\addlegendimage{only marks,very thick,dai_petrol, mark=square}
		\addlegendimage{only marks,very thick, magenta, mark=square}
		    \end{customlegend}
	\end{tikzpicture}};
		\node[anchor=west] (plot1) at (-10pt,210pt) {\input{tikz/data_entropy_correlation/NormalizedScnearioA1/plot.tex}};
    \node[black, anchor=west, rotate=90] at (-10pt,175pt) {{\footnotesize{Normalized Entropy [\%]}}};
	\node[black, anchor=west, rotate=90] at (260pt,175pt) {{\footnotesize{Normalized Entropy [\%]}}};
	\node[black, anchor=west] at (30pt,272pt) {{Asymetric Sensor Performance: Clear $\rightarrow$ Fog}};
	\node[black, anchor=west] at (285pt,272pt) {{Asymetric Sensor Performance: Day $\rightarrow$ Night}};

	\node[black, anchor=west] at (280pt,135 pt) {{\includegraphics[trim=0 0 0 0, clip, width=0.25\columnwidth]{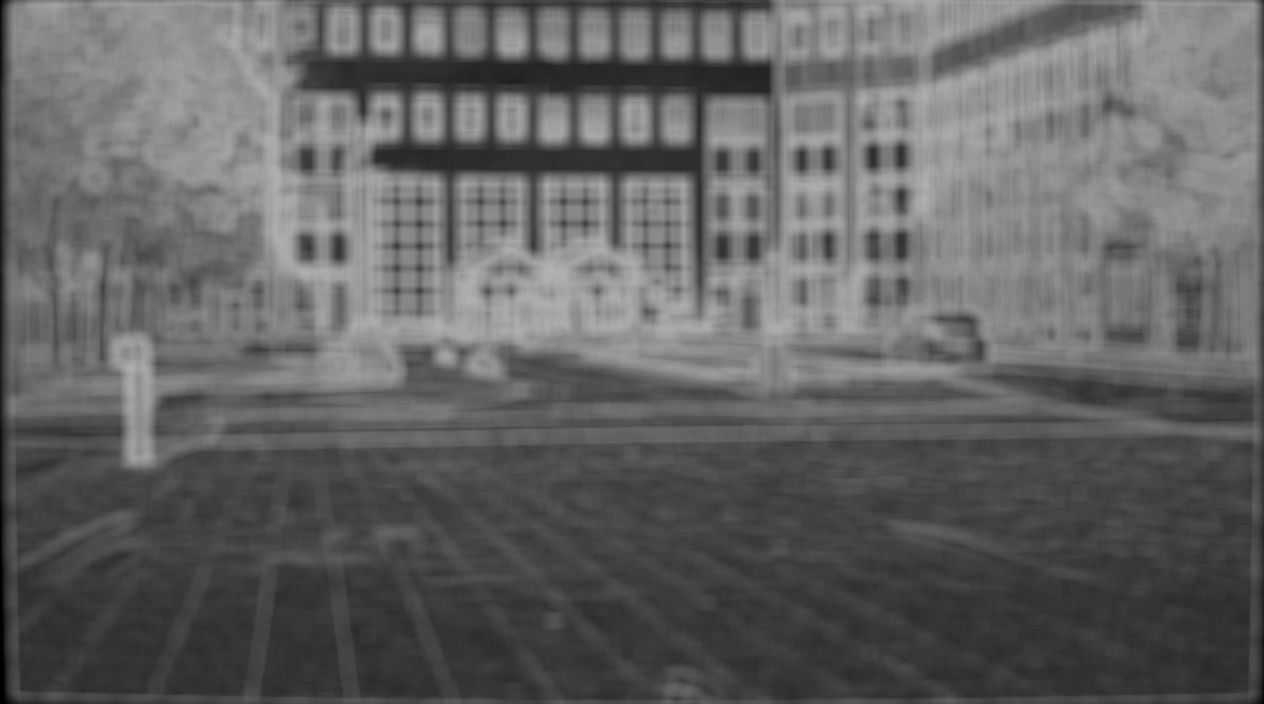}}};
	\node[black, anchor=west] at (280pt,100 pt) {{\includegraphics[trim=0 0 0 0, clip, width=0.25\columnwidth]{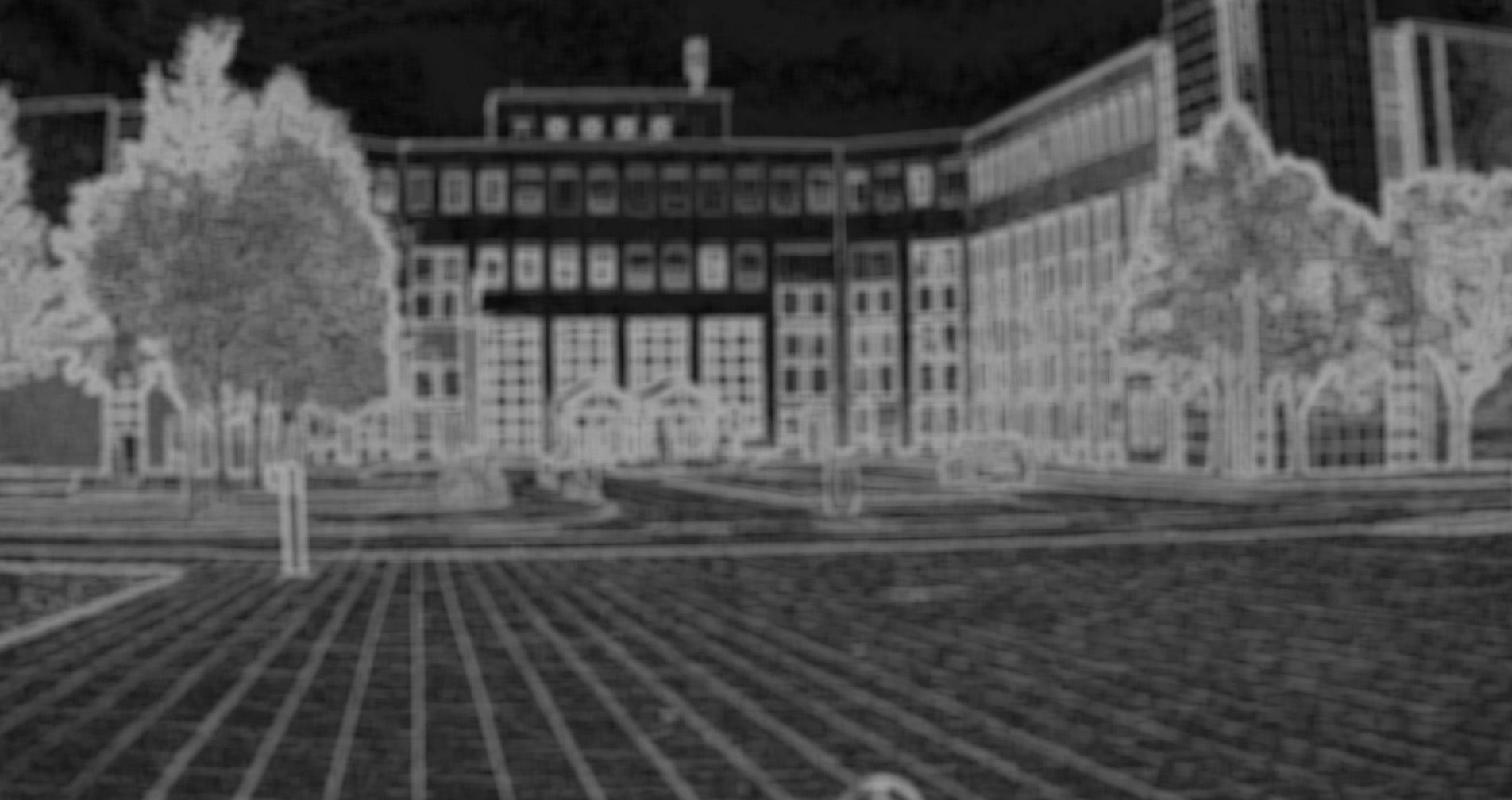}}};
	\node[black, anchor=west] at (280pt,65 pt) {{\includegraphics[trim=0 0 0 0, clip, width=0.25\columnwidth]{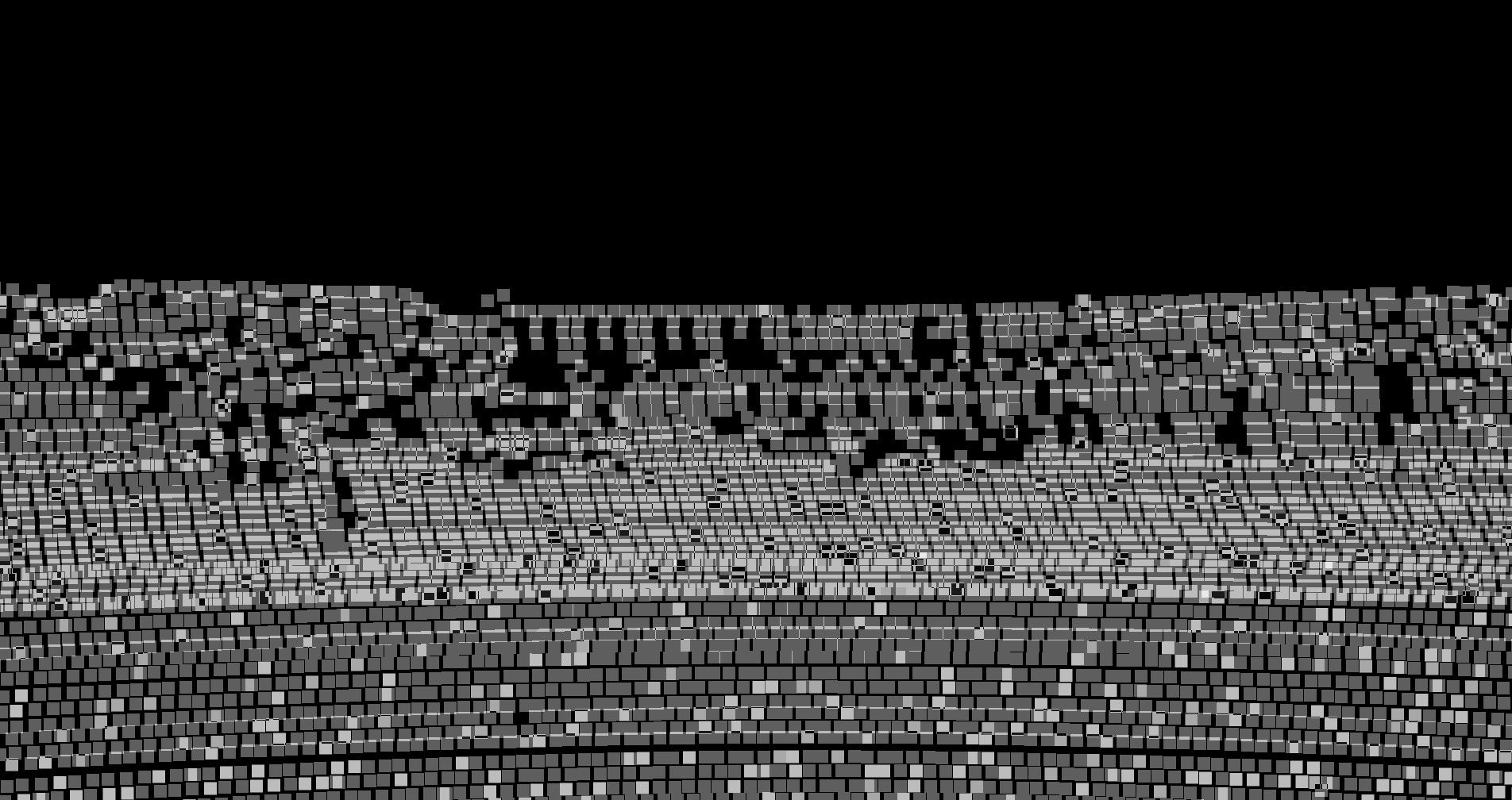}}};
	\node[black, anchor=west] at (280pt,30 pt) {{\includegraphics[trim=0 0 0 0, clip, width=0.25\columnwidth]{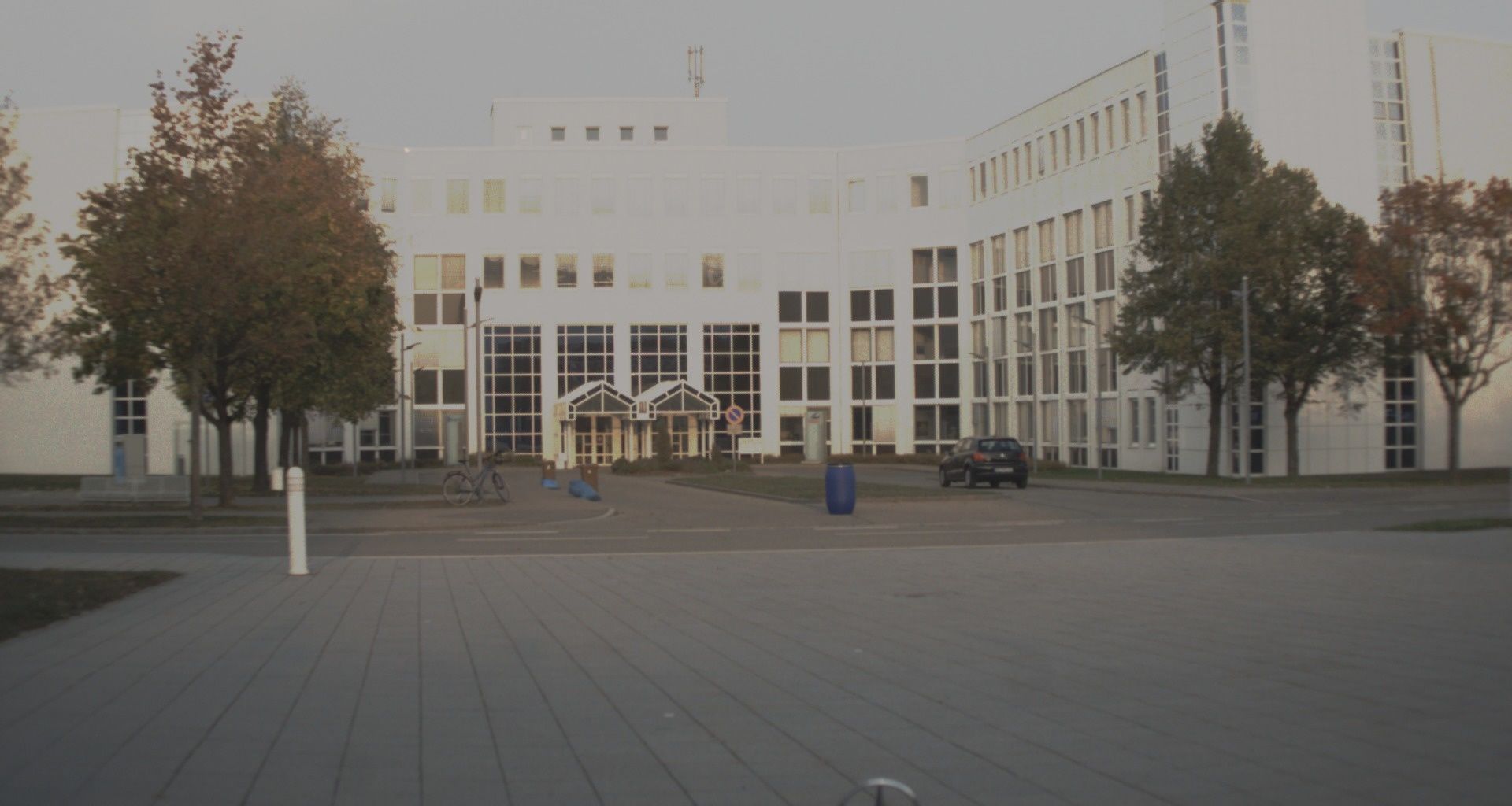}}};
	\node[mark size=3pt,color=dai_ligth_grey40K] at (290pt,145 pt) {\pgfuseplotmark{*}};
	\node[mark size=3pt,color=dai_deepred] at (287pt,110 pt) {\pgfuseplotmark{triangle*}};
	\node[mark size=3pt,color=dai_petrol] at (290pt,75 pt) {\pgfuseplotmark{square*}};
	\node[very thick, mark size=3pt,color=dai_deepred] at (287pt,40pt) {\pgfuseplotmark{triangle}};
		\draw [
		    thick,
		    decoration={
		        brace,
		    },
		    decorate
		] (281pt, 155pt) -- (344pt, 155pt);
	\draw[line width=0.5pt, densely dotted, -to]   (312.5pt,157pt) .. controls (312.5pt,170pt) and (285pt,157pt)  .. (285pt,184pt);
	\node[black, anchor=west] at (345pt,135 pt) {{\includegraphics[trim=0 0 0 0, clip, width=0.25\columnwidth]{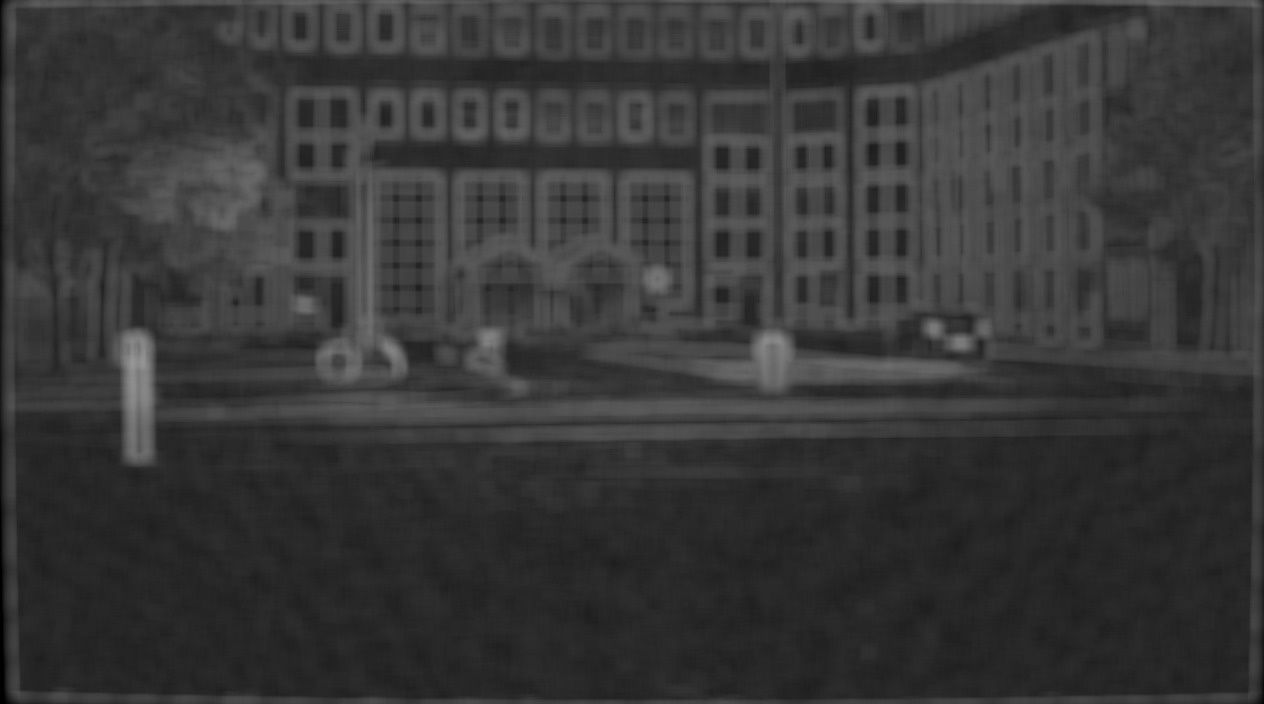}}};
	\node[black, anchor=west] at (345pt,100 pt) {{\includegraphics[trim=0 0 0 0, clip, width=0.25\columnwidth]{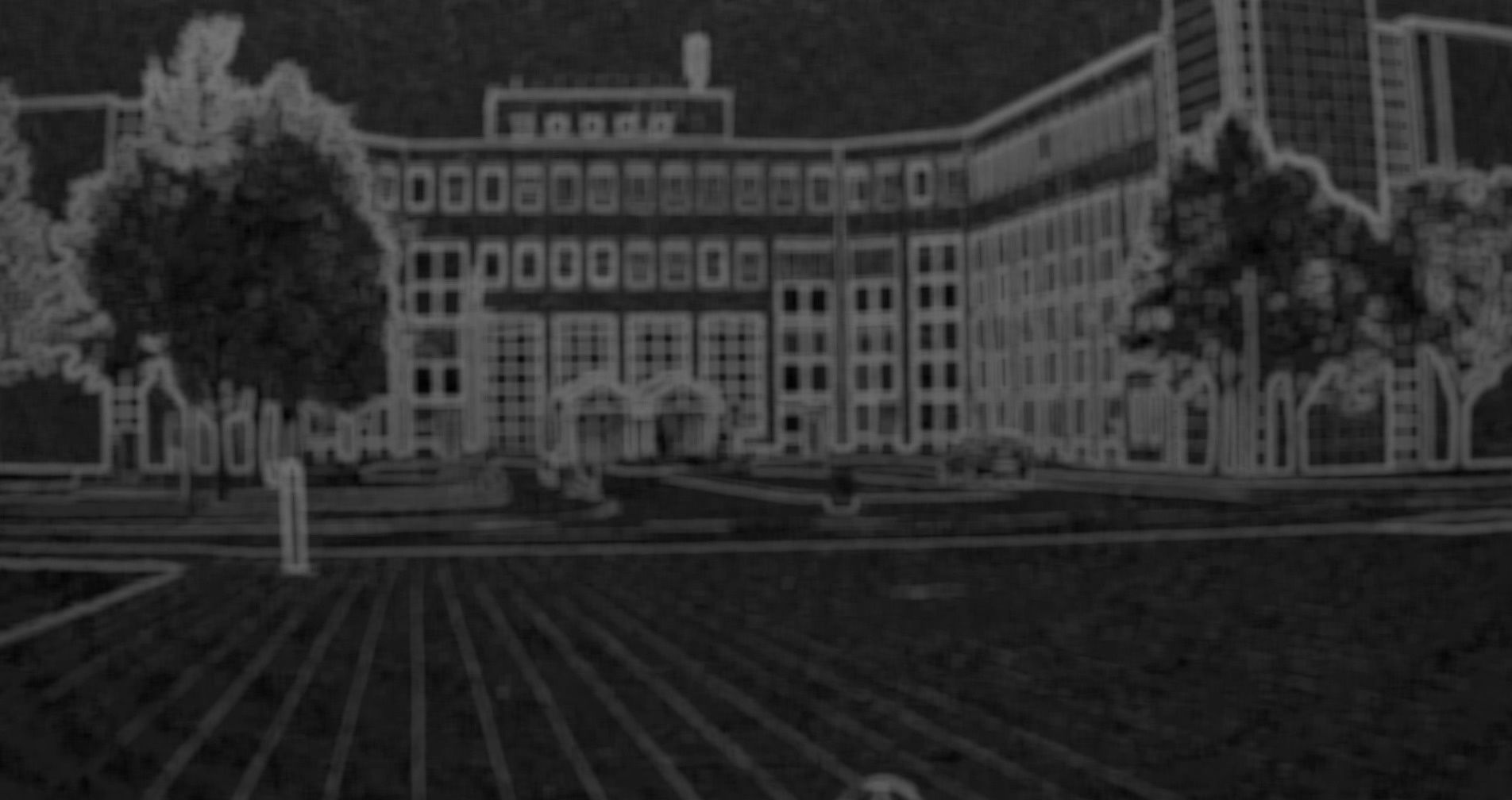}}};
	\node[black, anchor=west] at (345pt,65 pt) {{\includegraphics[trim=0 0 0 0, clip, width=0.25\columnwidth]{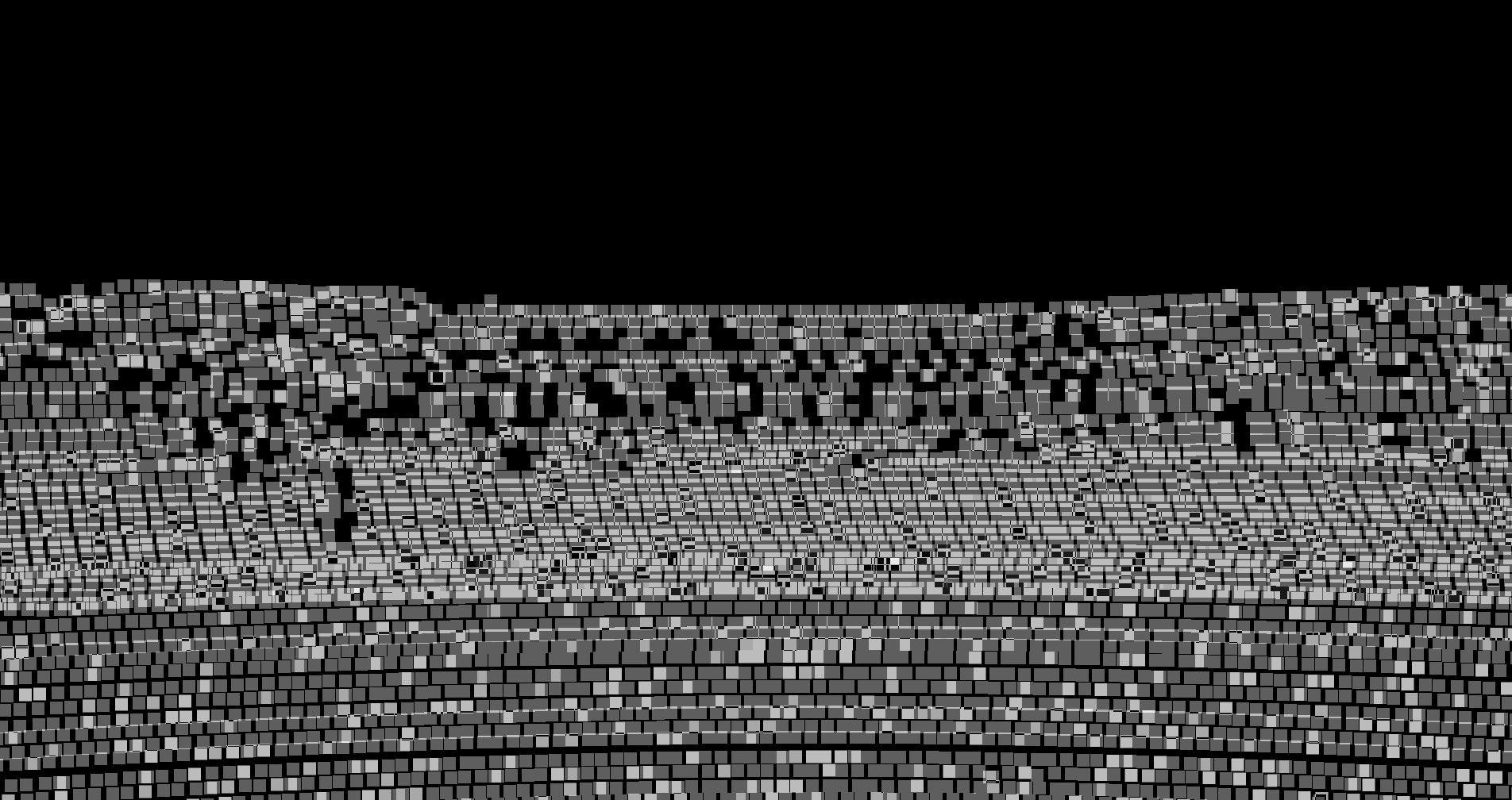}}};
	\node[black, anchor=west] at (345pt,30 pt) {{\includegraphics[trim=0 0 0 0, clip, width=0.25\columnwidth]{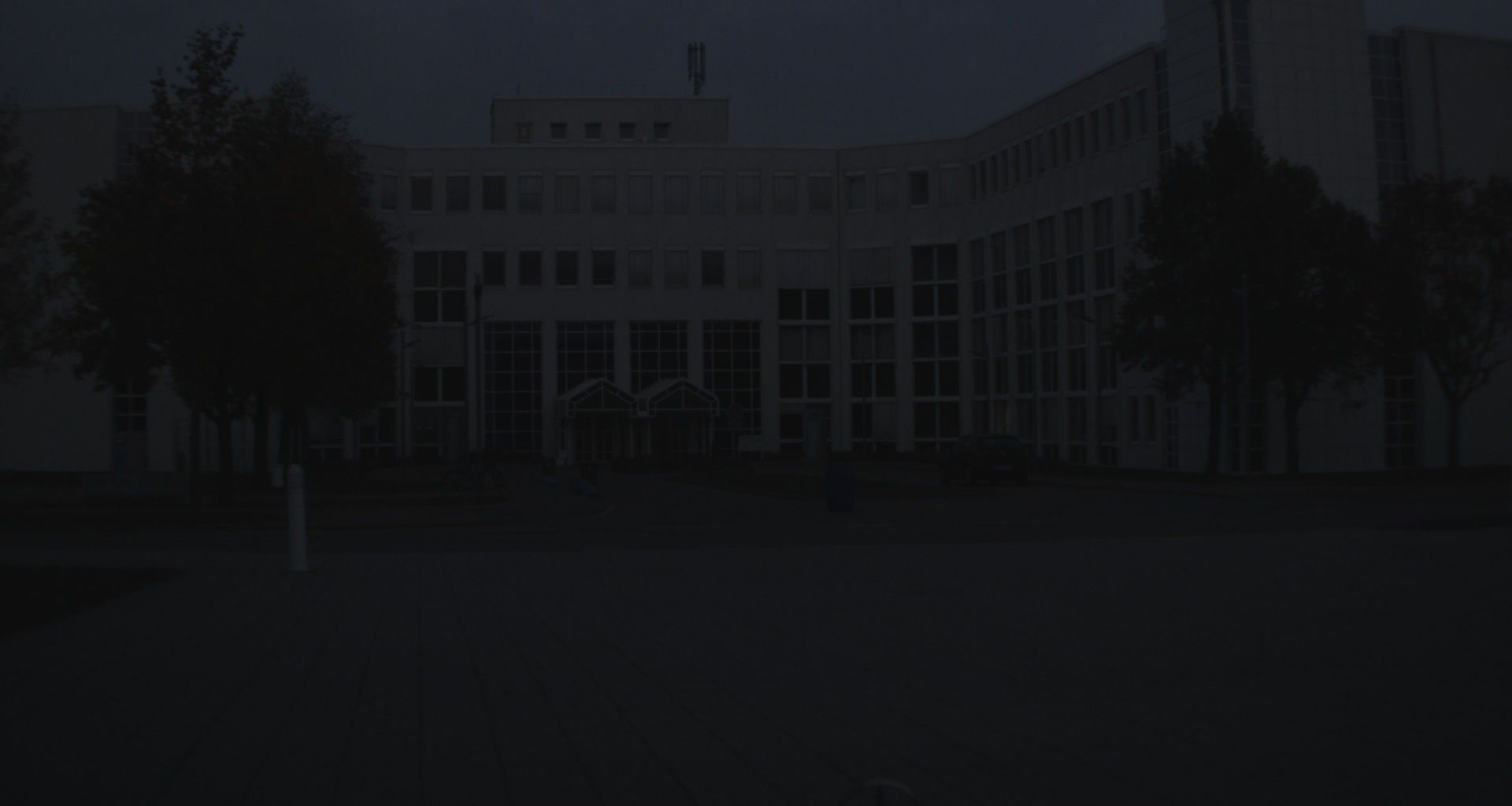}}};
	\node[mark size=3pt,color=dai_ligth_grey40K] at (355pt,145 pt) {\pgfuseplotmark{*}};
	\node[mark size=3pt,color=dai_deepred] at (352pt,110 pt) {\pgfuseplotmark{triangle*}};
	\node[mark size=3pt,color=dai_petrol] at (355pt,75 pt) {\pgfuseplotmark{square*}};
	\node[very thick, mark size=3pt,color=dai_deepred] at (352pt,40pt) {\pgfuseplotmark{triangle}};
		\draw [
		    thick,
		    decoration={
		        brace,
		    },
		    decorate
		] (346pt, 155pt) -- (409pt, 155pt);
	\draw[line width=0.5pt, densely dotted, -to]   (377.5pt,157pt) .. controls (377.5pt,170pt) and (348pt,157pt)  .. (348pt,184pt);
	\node[black, anchor=west] at (410pt,135 pt) {{\includegraphics[trim=0 0 0 0, clip, width=0.25\columnwidth]{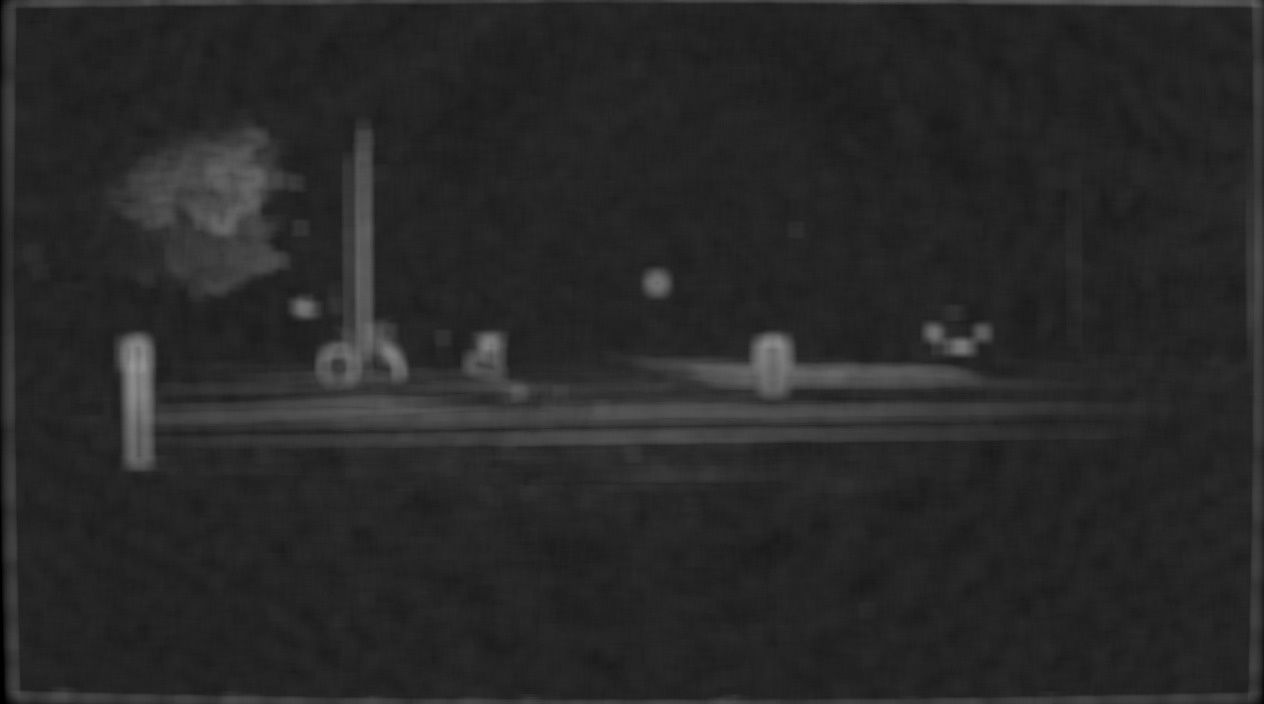}}};
	\node[black, anchor=west] at (410pt,100 pt) {{\includegraphics[trim=0 0 0 0, clip, width=0.25\columnwidth]{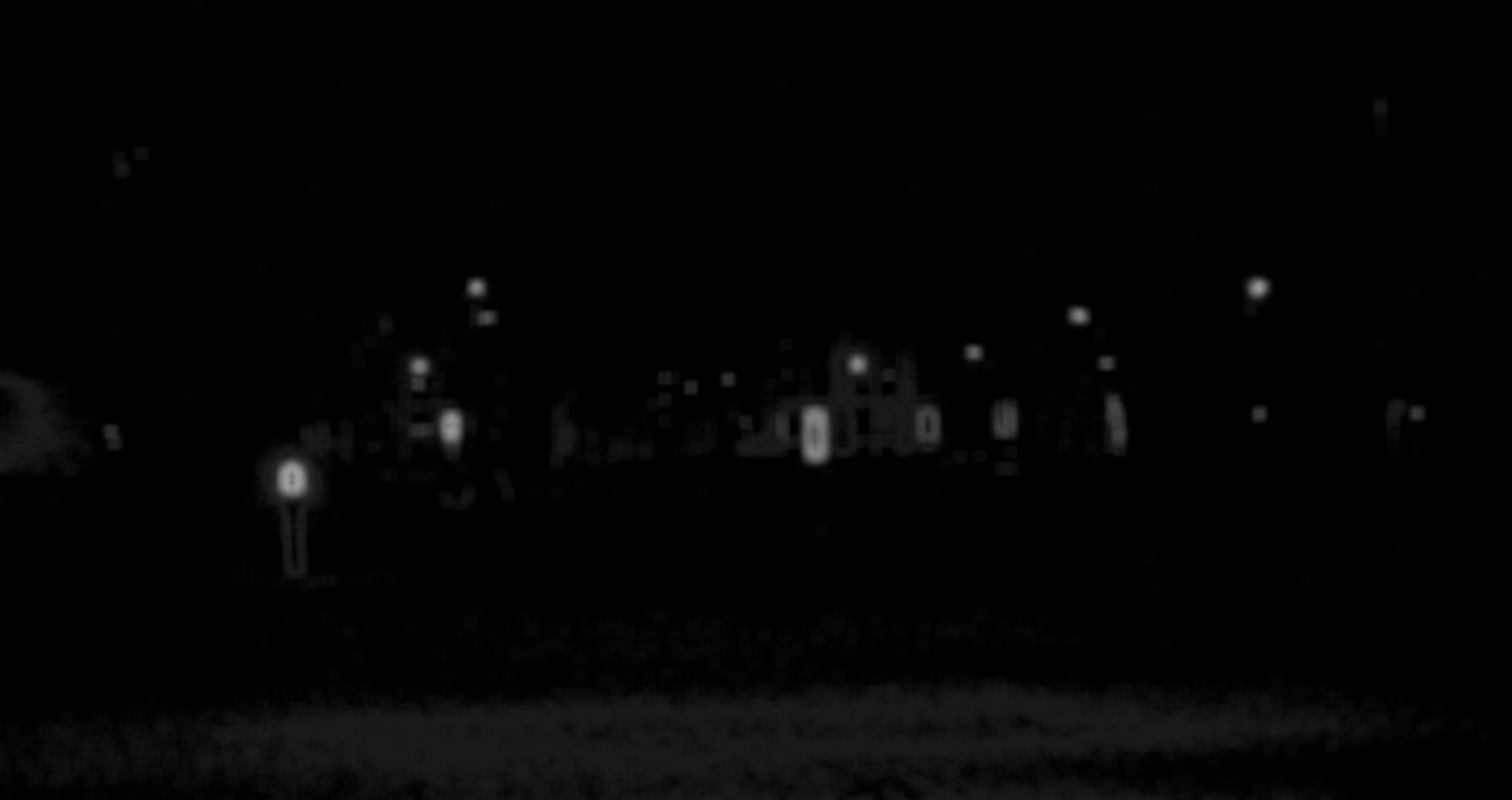}}};
	\node[black, anchor=west] at (410pt,65 pt) {{\includegraphics[trim=0 0 0 0, clip, width=0.25\columnwidth]{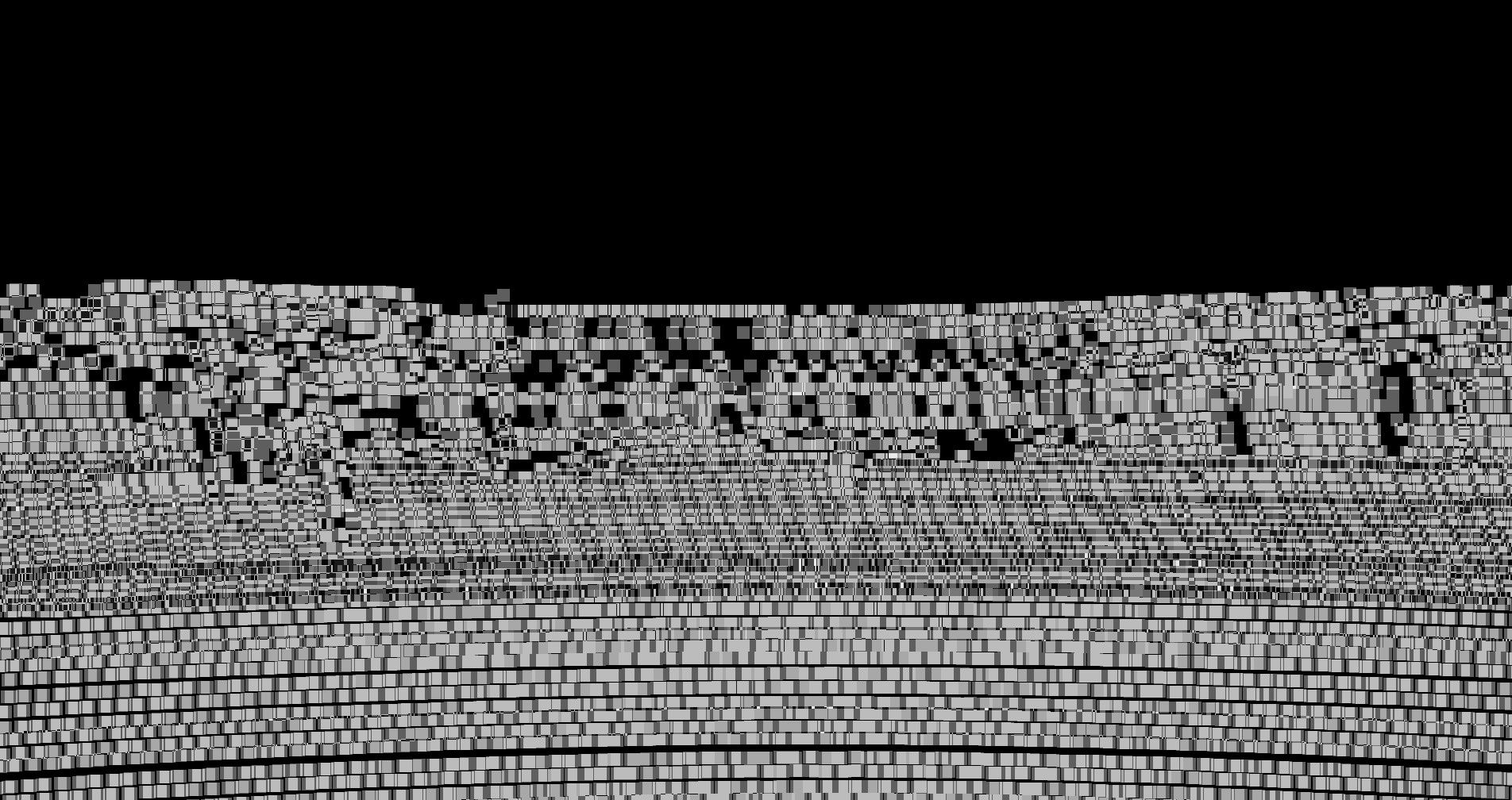}}};
	\node[black, anchor=west] at (410pt,30 pt) {{\includegraphics[trim=0 0 0 0, clip, width=0.25\columnwidth]{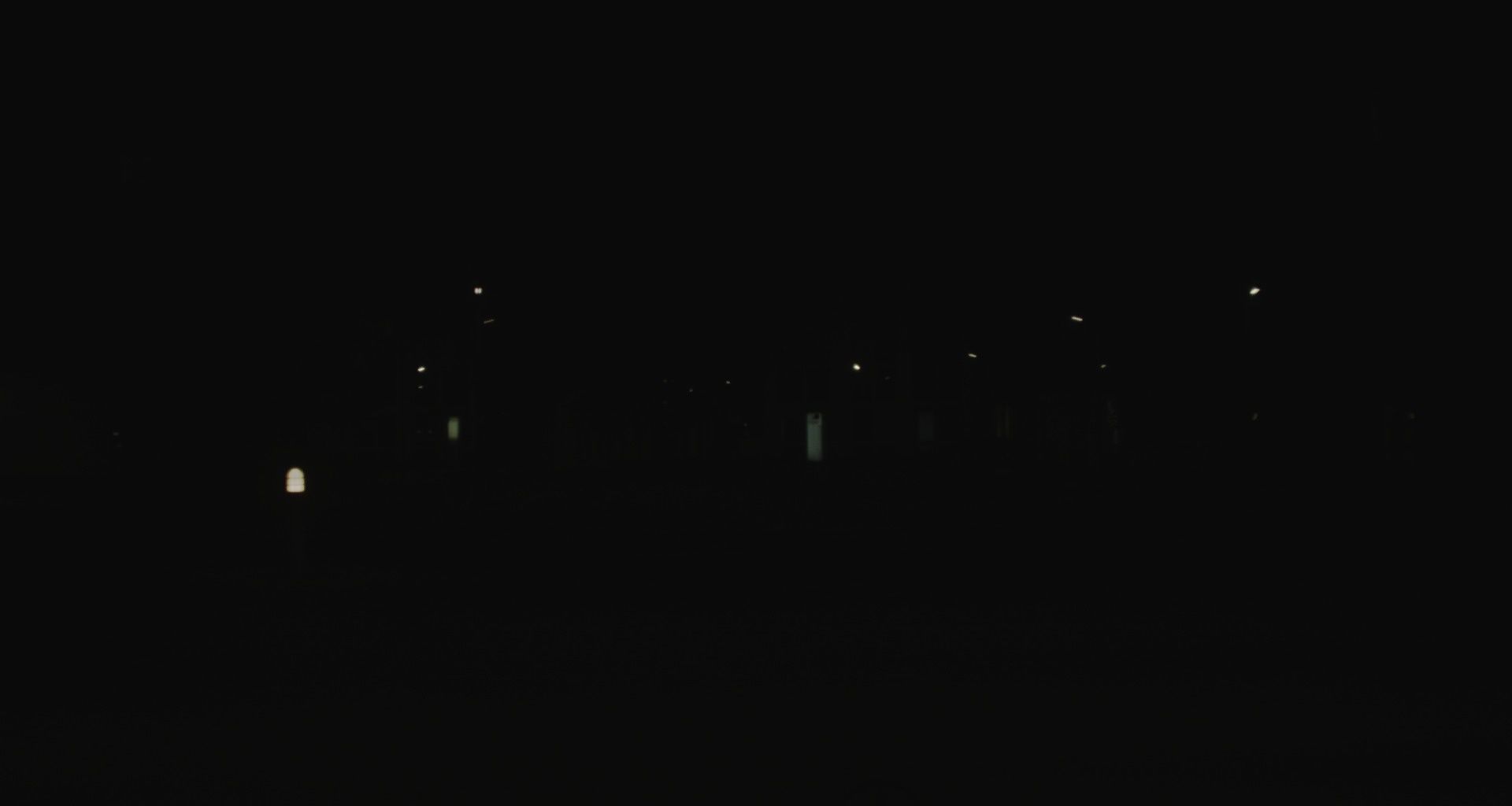}}};
	\node[mark size=3pt,color=dai_ligth_grey40K] at (420pt,145 pt) {\pgfuseplotmark{*}};
	\node[mark size=3pt,color=dai_deepred] at (417pt,110 pt) {\pgfuseplotmark{triangle*}};
	\node[mark size=3pt,color=dai_petrol] at (420pt,75 pt) {\pgfuseplotmark{square*}};
	\node[very thick, mark size=3pt,color=dai_deepred] at (417pt,40pt) {\pgfuseplotmark{triangle}};
		\draw [
		    thick,
		    decoration={
		        brace,
		        mirror,
		    },
		    decorate
		] (474pt, 155pt) -- (411pt, 155pt);
	\draw[line width=0.5pt, densely dotted, -to]   (442.5pt,157pt) .. controls (442.5pt,170pt) and (411.5pt,157pt)  .. (411.5pt,184pt);
	%
	\node[anchor=west] (plot2) at (260pt,210pt) {\input{tikz/data_entropy_correlation/DaytimeToNight/plot.tex}};
	\end{tikzpicture}
	\vspace{-2.5em}
	\caption{Normalized entropy with respect to the clear reference recording for a gated camera, RGB camera, radar, and lidar in varying fog visibilities (left) and changing illumination (right). The entropy has been calculated based on a dynamic scenario within a controlled fog chamber illustrated in Figure~\ref{fig:sensor_performance} (left) and a static scenario with changing natural illumination settings (right). Quantitative numbers have been calculated following Eq.~\eqref{eq:entropy}. Note the asymmetric sensor failure for different sensor technologies. Qualitative results are given below and are connected via arrows to their corresponding fog density/daytime.}\vspace{-1.5em}
	\label{fig:entropycorrelation}
\end{figure*}
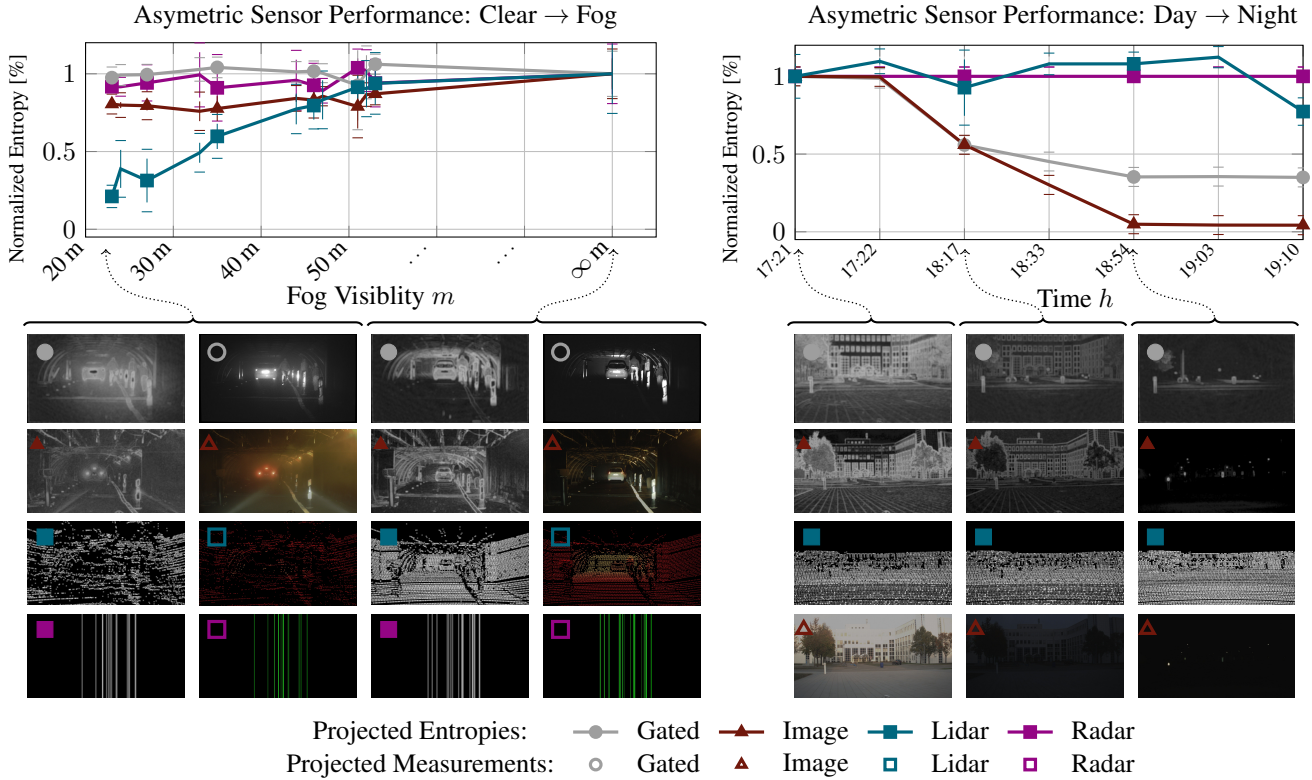

\subsection{Entropy-steered Fusion}\label{sec:entropy}
To steer the deep fusion towards redundant and reliable information, we introduce an entropy channel in each sensor stream, instead of directly inferring the adverse weather type and strength as in \cite{fogest2014,tarel2010improved}. We estimate local measurement entropy,
\vspace{-14pt}
\begin{equation}\label{eq:entropy}
\begin{aligned}
\rho & = \sum_{m,n}^{w,h}\sum_{i=0}^{255} p_i^{mn} \log\left(p_i^{mn}\right) \;\;\; \text{,with} \\\vspace{-0.3em}
p_i^{mn} & =\frac{1}{MN}\sum_{j,k}^{M,N} \delta\left(I(m+j,n+k)-i\right).
\end{aligned}
\vspace{-1.3em}
\end{equation}
The entropy is calculated for each \unit[8]{bit} binarized stream $I$ with pixel values $i\ \epsilon\left[0,255\right]$ in the proposed image-space data representation. Each stream is split into patches of size $M \times N = \unit[16]{px} \times \unit[16]{px}$ resulting in a $w \times h = \unit[1920]{px}\times\unit[1024]{px}$ entropy map. The multimodal entropy maps for two different scenarios are shown in Figure~\ref{fig:entropycorrelation}: the left scenario shows a scene containing a vehicle, cyclist, and pedestrians in a controlled fog chamber. The passive RGB camera and lidar suffer from backscatter and attenuation with decreasing fog visibilities, while the gated camera suppresses backscatter through gating. Radar measurements are also not substantially degraded in fog. The right scenario in Figure~\ref{fig:entropycorrelation} shows a static outdoor scene under varying ambient lighting. In this scenario, active lidar and radar are not affected by changes in ambient illumination. For the gated camera, the ambient illumination disappears, leaving only the actively illuminated areas, while the passive RGB camera degenerates with decreasing ambient light. 

The steering process is learned purely on clean weather data, which contains different illumination settings present in day to night-time conditions. No real adverse weather patterns are presented during training. Further, we drop sensor streams randomly with probability 0.5 and set the entropy to a constant zero value. 

\subsection{Loss Functions and Training Details}\label{sec:training}
The number of anchor boxes in different feature layers and their sizes play an important role during training and are given in the supplemental material. In total, each anchor box with class label $y_i$ and probability $p_i$ is trained using the cross entropy loss with softmax, 
\vspace{-5pt}
\begin{equation}
H(p)=\sum_i (y_i\log(p_i)+(1 - y_i)\log(1 - p_i)).
\end{equation}
The loss is split up for positive and negative anchor boxes with a matching threshold of $0.5$. For each positive anchor box, the bounding box coordinates $x$ are regressed using a Huber loss $H(x)$ given by,
\vspace{-10pt}
\begin{equation}
H(x)=\left\lbrace
\begin{array}{cc}
x^2 / 2, &  \mathrm{if} \left|x\right| < 1\\
\left|x\right| - 0.5, &    \mathrm{if} \left|x\right| > 1
\end{array}\right.\vspace{-4pt}
\end{equation}
The total number of negative anchors is restricted to $5 \times$ the number of positive examples using hard example mining \cite{SSDLiu2015,Shrivastava16}. All networks are trained from scratch with a constant learning rate and L2 weight decay of 0.0005. 

%% file: tikz/data_entropy_correlation/NormalizedScnearioA1/plot.tex
	\begin{tikzpicture}
	\begin{axis}[
	axis background/.style={fill=white},
	xlabel=Fog Visiblity $m$,
    xmin=20,
	xmax=85,
	ymin=-0.05,
	ymax=1.2,    
	grid=major,
	width=1.1\columnwidth,
	height=0.5\columnwidth,
	x tick label style={rotate=45, anchor=east, align=center, font=\small},
	xtick={20,30,40,50,60,70,80},
	xticklabels={\unit[20]{m},\unit[30]{m},\unit[40]{m},\unit[50]{m},\scriptsize{$\cdots$},\scriptsize{$\cdots$}, \unit[$\infty$]{m}},
	]

	\addplot+ [very thick, solid, color=magenta,
	mark=square*,
	mark size={2}, 
	mark options={solid,magenta},  
	mark repeat={2},
	error bars/.cd,
	y dir=both,
	y explicit, 
	error mark options={rotate=90,magenta}
	]
	table[col sep=comma, x={x}, y={mean}, y error = {std}]{tikz/data_entropy_correlation/NormalizedScnearioA1/radar_entropy.txt};
		
	\addplot+ [very thick, solid, dai_ligth_grey40K,mark=*,
	mark size={2},
	mark options={solid,dai_ligth_grey40K}, 
	mark repeat={2},
	error bars/.cd,
	y dir=both,
	y explicit, 
	error mark options={rotate=90,dai_ligth_grey40K}
	]
	table[col sep=comma, x={x}, y={mean}, y error = {std}]{tikz/data_entropy_correlation/NormalizedScnearioA1/gated_entropy.txt};
	
	\addplot+ [very thick, solid, color=dai_deepred,
	mark=triangle*,
	mark size={2}, 
	mark options={solid,dai_deepred}, 
	mark repeat={2},
	error bars/.cd,
	y dir=both,
	y explicit, 
	error mark options={rotate=90,dai_deepred}
	]
	table[col sep=comma, x={x}, y={mean}, y error = {std}]{tikz/data_entropy_correlation/NormalizedScnearioA1/image_entropy.txt};
	\addplot+ [very thick, solid, color=dai_petrol,
	mark=square*,
	mark size={2}, 
	mark options={solid,dai_petrol},  
	mark repeat={2},
	error bars/.cd,
	y dir=both,
	y explicit, 
	error mark options={rotate=90,dai_petrol}
	]
	table[col sep=comma, x={x}, y={mean}, y error = {std}]{tikz/data_entropy_correlation/NormalizedScnearioA1/lidar_entropy.txt};
	\end{axis}
	\end{tikzpicture}

%% file: tikz/data_entropy_correlation/DaytimeToNight/plot.tex
	\begin{tikzpicture}
	\begin{axis}[
	axis background/.style={fill=white},
	xlabel=\qquad Time $h$,
	xmin=1,
	xmax=7,
	ymin=-0.05,
	ymax=1.2,  
	grid=major,
	width=\columnwidth,
	height=0.5\columnwidth,
	x tick label style={rotate=45, anchor=east, align=center, font=\small},
	xtick={0,1,2,3,4,5,6,7,8},
	xticklabels={\scriptsize{17:05},\scriptsize{17:21},\scriptsize{17:22},\scriptsize{18:17},\scriptsize{18:33},\scriptsize{18:54},\scriptsize{19:03}, \scriptsize{19:10}, \scriptsize{19:21}},
	]
	
%
	
	\addplot+ [very thick, solid, color=magenta,
	mark=square*,
	mark size={2}, 
	mark options={solid,magenta},  
	mark repeat={2},
	error bars/.cd,
	y dir=both,
	y explicit, 
	error mark options={rotate=90,magenta}
	]
	table[col sep=comma, x={x}, y={mean}, y error = {std}]{tikz/data_entropy_correlation/DaytimeToNight/radar_entropy_day_night2.txt};
		
	\addplot+ [very thick, solid, dai_ligth_grey40K,mark=*,
	mark size={2},
	mark options={solid,dai_ligth_grey40K}, 
	mark repeat={2},
	error bars/.cd,
	y dir=both,
	y explicit, 
	error mark options={rotate=90,dai_ligth_grey40K}
	]
	table[col sep=comma, x={x}, y={mean}, y error = {std}]{tikz/data_entropy_correlation/DaytimeToNight/gated_entropy_day_night2.txt};
	
	\addplot+ [very thick, solid, color=dai_deepred,
	mark=triangle*,
	mark size={2}, 
	mark options={solid,dai_deepred}, 
	mark repeat={2},
	error bars/.cd,
	y dir=both,
	y explicit, 
	error mark options={rotate=90,dai_deepred}
	]
	table[col sep=comma, x={x}, y={mean}, y error = {std}]{tikz/data_entropy_correlation/DaytimeToNight/image_entropy_day_night2.txt};
	\addplot+ [very thick, solid, color=dai_petrol,
	mark=square*,
	mark size={2}, 
	mark options={solid,dai_petrol},  
	mark repeat={2},
	error bars/.cd,
	y dir=both,
	y explicit, 
	error mark options={rotate=90,dai_petrol}
	]
	table[col sep=comma, x={x}, y={mean}, y error = {std}]{tikz/data_entropy_correlation/DaytimeToNight/lidar_entropy_day_night2.txt};
%
	
%
%
	
	\end{axis}
	\end{tikzpicture}

%% file: result.tex
\begin{table*}[t]
\vspace{-0.3em}
    \footnotesize
    \heavyrulewidth 1.7pt
    \belowrulesep 2pt
    \aboverulesep 2pt
    \centering
    \resizebox{.99\linewidth}{!}{
        \begin{tabular}{@{}l|@{\hskip 2\tabcolsep}ccc@{\hskip 2\tabcolsep}|@{\hskip 2\tabcolsep}ccc@{\hskip 2\tabcolsep}|@{\hskip 2\tabcolsep}ccc@{\hskip 2\tabcolsep}@{}|@{\hskip 2\tabcolsep}ccc@{\hskip 0\tabcolsep}@{}}
		\toprule
		\textsc{\textbf{Weather}}       & & \textbf{clear} &  &  & \textbf{light fog} &   &  &\textbf{dense fog}& & &\textbf{snow/rain}&\\ 
		\textsc{Difficulty}            		   &  easy  & mod. & hard & easy  & mod.  & hard   & easy  & mod.  & hard & easy  & mod.  & hard \\
		\toprule
       \textsc{\textbf{Deep Entropy Fusion (this work)}}  & 89.84 & \textbf{85.57} & \textbf{79.46} & {90.54} &  \textbf{87.99} & \textbf{84.90} & {87.68} & \textbf{81.49} & \textbf{76.69} & {88.99} & \textbf{83.71} & \textbf{77.85}  \\   
		\textsc{\textbf{Deep Fusion (this work)}} &  \textbf{90.07} & 80.31 & 77.82 &  \textbf{90.60} & 81.08 & 79.63 & 86.77 & 77.28 & 73.93 & \textbf{89.25} & 79.09 & 70.51\\
        \textsc{Fusion SSD}	& 87.73 & 78.02 & 69.49 & 88.33 & 78.65 & 76.54 & 74.07 & 68.46 & 63.23 & 85.49 & 75.28 & 67.48 \\
        \textsc{Concat. SSD}	& 86.12 & 76.62 & 68.61 & 87.98 & 78.24 & 70.17 & 77.99 & 69.16 & 67.07 & 83.63 & 73.65 & 66.26   \\
	    \textsc{ADDA~\cite{ADDA}\footnotemark[1]} & 85.27 & 70.51 & 67.86 & 87.83 & 78.68 & 70.38 & {87.64} & 78.12 & 74.37 & 84.17 & 74.25 & 66.86 \\
		\textsc{CyCada~\cite{Cycada}\footnotemark[1]}	& 88.50 & 77.84 & 69.56 & 89.08 & 79.36 & 75.58 & 87.24 & 77.04 & 73.38 & 85.56 & 74.80 & 67.22\\
        \textsc{Image-only SSD} & 85.43 & 75.75 & 67.79 & 87.76 & 78.52 & 70.43 & \textbf{87.89} & 78.25 & 74.96 & 84.33 & 74.38 & 67.01  \\
        \textsc{Gated-only SSD} & 77.10 & 61.95 & 58.27 & 80.65 & 69.64 & 61.75 & 75.16 & 66.76 & 61.68 & 77.32 & 61.31 & 57.23 \\
        \textsc{Lidar-only SSD} & 73.46 & 57.32 & 54.62 & 68.43 & 54.82 & 51.91 & 28.98 & 25.24 & 24.56 & 67.50 & 52.26 & 46.83 \\
        \textsc{Radar-only SSD} & 10.26 & 8.54 & 8.23 & 16.92 & 13.24 & 12.66 & 16.33 & 13.57 & 13.00 & 12.94 & 10.95 & 10.40 \\
		\textsc{AVOD-FPN \cite{ku2018joint}}	& 66.47 & 58.71 & 51.63 & 60.40 & 52.51 & 51.92 & 33.95 & 26.29 & 26.17 & 59.55 & 51.91 &  50.54\\
		\textsc{Frustum PointNet \cite{Qi2017}}  & 80.06 &  75.89 & 67.70 & 84.06 &  76.88 & 75.44 & 76.69 & 73.62 & 68.49 & 78.34 & 74.34 & 66.52\\
     	\bottomrule
		\end{tabular}}
	\vspace{-1.0em}
    \caption{Quantitative detection AP on real unseen weather-affected data from dataset split across weather and difficulties easy/moderate/hard following~\cite{Kitt_dataset}. All detection models except domain adaptation approaches are trained solely on clean data without weather distortions. The best model is highlighted in bold. }\label{tab:TestResults} \vspace{-1.7em}
\end{table*}
In this section, we validate the proposed fusion model on unseen experimental test data. We compare the method against existing detectors for single sensory inputs and fusion methods, as well as domain adaptation methods. Due to the weather-bias of training data acquisition, we only use the clear weather portion of the proposed dataset for training. We assess the detection performance using our novel multimodal weather dataset as a test set, see supplemental data for test and training split details. 

\vspace{0.3em}
We validate the proposed approach in Table~\ref{tab:TestResults}, which we dub Deep Entropy Fusion, on real adverse weather data. We report Average Precision (AP) for three different difficulty levels (easy, moderate, hard) and evaluate on cars following the KITTI evaluation framework \cite{Kitt_dataset} at various fog densities, snow disturbances, and clear weather conditions. We compare the proposed model against recent state-of-the-art lidar-camera fusion models, including AVOD-FPN~\cite{ku2018joint}, Frustum PointNets~\cite{Qi2017}, and variants of the proposed method with alternative fusion or sensory inputs. As baseline variants, we implement two fusion and four single sensor detectors. In particular, we compare against late fusion with image, lidar, gated, and radar features concatenated just before bounding-box regression (\emph{Fusion SSD}), and early fusion by concatenating all sensory data at the early beginning of one feature extraction stack (\emph{Concat SSD}). The \emph{Fusion SSD} network shares the same structure as the proposed model, but without the feature exchange and the adaptive fusion layer.  Moreover, we compare the proposed model against an identical SSD branch with single sensory input (\emph{Image-only SSD}, \emph{Gated-only SSD}, \emph{Lidar-only SSD}, \emph{Radar-only SSD}). All models were trained with identical hyper-parameters and anchors.

Evaluated on adverse weather scenarios, the detection performance decrease for all methods. Note that assessment metrics can increase simultaneously as scene complexity changes between the weather splits. For example, when fewer vehicles participate in road traffic or the distance between vehicles increases in {icy} conditions, fewer vehicles are occluded. While the performance for image and gated data is almost steady, it decreases substantially for lidar data while it increases for radar data. The decrease in lidar performance can be described by the strong backscatter, see Supplemental Material. As a maximum of 100 measurement targets limits the performance of the radar input, the reported improvements are resulting from simpler scenes.

\footnotetext{Requires large corpus of adverse weather data for training.}

Overall, the large reduction in lidar performance for foggy conditions affects the lidar only detection rate by a drop in 45.38\,\% AP. Furthermore, it also has a strong impact on camera-lidar fusion models \emph{AVOD}, \emph{Concat SSD} and \emph{Fusion SSD}. Learned redundancies no longer hold, and these methods even fall below image-only methods.

Two-stage methods, such as Frustum PointNet~\cite{Qi2017}, drop {quickly}. However, they asymptotically achieve higher results compared to AVOD, because the statistical priors learned for the first stage are based on \emph{Image-only SSD} that limits its performance to image-domain priors. AVOD is limited by several assumptions that hold for clear weather, such as the importance sampling of boxes filled with lidar data during training, achieving the lowest fusion performance overall.  Moreover, as the fog density increases, the proposed adaptive fusion model outperforms all other methods. Especially under severe distortions, the proposed adaptive fusion layer results in significant margins over the model without it (\emph{Deep Fusion}). Overall the proposed method outperforms all baseline approaches. In dense fog, it improves by a margin of 9.69\,\% compared to the next-best feature-fusion variant. 

For completeness, we also compare the proposed model to recent domain adaptation methods. First, we adapt our \emph{Image-Only SSD} features from clear weather to adverse weather following \cite{ADDA}. Second, we investigate the style transfer from clear weather to adverse weather utilizing \cite{Cycada} and generate adverse weather training samples from clear weather input. Note that these methods have an unfair advantage over all other compared approaches as they have seen adverse weather scenarios sampled from our validation set. Note that domain adaptation methods cannot be directly applied as they need target images from a specific domain. Therefore, they do also not offer a solution for rare edge cases with limited data. Furthermore \cite{Cycada} \emph{does not model distortions}, including fog or snow, see experiments in the Supplemental Material. We note that synthetic data augmentation following \cite{sakaridis2018semantic} or image-to-image reconstruction methods that remove adverse weather effects~\cite{wang2018pix2pixHD} do neither affect the reported margins of the proposed multimodal deep entropy fusion.

%% file: discussion.tex
In this paper we address a critical problem in autonomous driving: multi-sensor fusion in scenarios, where annotated data is sparse and difficult to obtain due to natural weather bias. To assess multimodal fusion in adverse weather, we introduce a novel adverse weather dataset covering camera, lidar, radar, gated NIR, and FIR sensor data. The dataset contains rare scenarios, such as heavy fog, heavy snow, and severe rain, during more than \unit[10,000]{km} of driving in northern Europe.
We propose a real-time deep multimodal fusion network which departs from proposal-level fusion, and instead adaptively fuses driven by measurement entropy. Exciting directions for future research include the development of end-to-end models enabling the failure detection and an adaptive sensor control as noise level or power level control in lidar sensors.